\documentclass[10pt,journal,compsoc]{IEEEtran}
\ifCLASSOPTIONcompsoc
\usepackage[nocompress]{cite}
\else
\usepackage{cite}
\fi
\usepackage{algorithm}
\usepackage{algorithmic}
\usepackage{setspace}
\usepackage{amsmath}
\usepackage{amsfonts}
\usepackage{CJK}
\usepackage{indentfirst}
\usepackage{bm}
\usepackage{graphicx}
\makeatletter
\let\MYcaption\@makecaption
\makeatother

\usepackage[font=footnotesize]{subcaption}

\makeatletter
\let\@makecaption\MYcaption
\makeatother

\usepackage[hidelinks,colorlinks=true,linkcolor=blue,citecolor=blue]{hyperref}
\usepackage{booktabs}
\usepackage{multirow}
\usepackage{amssymb}
\usepackage{color}

\usepackage{float}
\usepackage{multicol}
\usepackage{amsthm}

\usepackage{cite}

\ifCLASSINFOpdf
\else
\fi

\ifCLASSINFOpdf
\else
\fi
\begin{document}
		%
		\title{Degradation-Aware Residual-Conditioned Optimal Transport for Unified Image Restoration}
		%
		%
		%
		%
		
		\author{Xiaole Tang,
			Xiang Gu,
			Xiaoyi He,
			Xin Hu,
			Jian Sun
			\IEEEcompsocitemizethanks{
				\IEEEcompsocthanksitem The authors are with School of Mathematics and Statistics, Xi'an Jiaotong University, Shaanxi, P.R. China. E-mail: \{tangxl, hexiaoyi, huxin7020,\}@stu.xjtu.edu.cn,  \{xianggu, jiansun\}@xjtu.edu.cn.
			}
			\thanks{(Corresponding author: Jian Sun.)
		}}
		
		%
		%

	\markboth{Journal of \LaTeX\ Class Files,~Vol.~14, No.~8, August~2015}%
	{Shell \MakeLowercase{\textit{et al.}}: Bare Demo of IEEEtran.cls for Computer Society Journals}
	%



	\IEEEtitleabstractindextext{%
		\begin{abstract}
			Unified, or more formally, all-in-one image restoration has emerged as a practical and promising low-level vision task for real-world applications. In this context, the key issue lies in how to deal with different types of degraded images simultaneously. Existing methods fit joint regression models over multi-domain degraded-clean image pairs of different degradations. However, due to the severe ill-posedness of inverting heterogeneous degradations, they often struggle with thoroughly perceiving the degradation semantics and rely on paired data for supervised training, yielding suboptimal restoration maps with structurally compromised results and lacking practicality for real-world or unpaired data. To break the barriers, we present a Degradation-Aware Residual-Conditioned Optimal Transport (DA-RCOT) approach that models (all-in-one) image restoration as an optimal transport (OT) problem for unpaired and paired settings, introducing the transport residual as a degradation-specific cue for both the transport cost and the transport map. Specifically, we formalize image restoration with a residual-guided OT objective by exploiting the degradation-specific patterns of the Fourier residual in the transport cost. More crucially, we design the transport map for restoration as a two-pass DA-RCOT map, in which the transport residual is computed in the first pass and then encoded as multi-scale residual embeddings to condition the second-pass restoration. This conditioning process injects intrinsic degradation knowledge (e.g., degradation type and level) and structural information from the multi-scale residual embeddings into the OT map, which thereby can dynamically adjust its behaviors for all-in-one restoration. Extensive experiments across five degradations demonstrate the favorable performance of DA-RCOT as compared to state-of-the-art methods, in terms of distortion measures, perceptual quality, and image structure preservation. Notably, DA-RCOT delivers superior adaptability to real-world scenarios even with multiple degradations and shows distinctive robustness to both degradation levels and the number of degradations. Source code is publicly available at: \url{https://github.com/xl-tang3/DA-RCOT}.
		\end{abstract}
		
		}

	\maketitle

	\IEEEdisplaynontitleabstractindextext

	\ifCLASSOPTIONpeerreview
	\begin{center} \bfseries EDICS Category: 3-BBND \end{center}
	\fi
	%
	\IEEEpeerreviewmaketitle

	\IEEEraisesectionheading{\section{Introduction}
		\label{sec:introduction}}
	\IEEEPARstart{I}{mage} restoration (IR)  holds a fundamental 
	position in low-level computer vision, aiming to address the degradation (e.g., most commonly noise, blur, rain, haze, low light) in a degraded image.
	Traditional methods focus on designing an optimization problem that exploits priors of natural image \cite{he2010single}, \cite{pan2016blind}, \cite{ulyanov2018deep}, \cite{xu2010image},  \cite{wang2021multi} or corresponding image-induced signals such as gradients \cite{fergus2006removing}, \cite{sun2008image} and residuals or noise maps \cite{shan2008high}, \cite{ji2011robust}, \cite{Tang_2023_CVPR}. Recent advances in deep learning techniques \cite{he2016deep}, \cite{vaswani2017attention}, \cite{dosovitskiy2021an}, \cite{liu2021swin} have triggered great achievements in image restoration, in which most methods \cite{lai2017deep}, \cite{lai2018fast}, \cite{zhang2018ffdnet}, \cite{Zamir2021MPRNet}, \cite{Zamir2021Restormer}, \cite{liang2021swinir}, \cite{chen2022simple} \cite{luo2023image}, \cite{zhou2023fourmer} train task-specific restoration models over degraded-clean image pairs of a single known degradation. However, this specificity limits their practicability in real-world applications, e.g., autonomous navigation \cite{levinson2011towards}, \cite{prakash2021multi}, surveillance systems \cite{liang2018deep}, and digital photography \cite{petschnigg2004digital}, where varied and unexpected degradations usually occur. Therefore, there is an emerging field known as ``All-in-One" image restoration (AIR) that aims to address different degradations simultaneously.

\begin{figure}[!t]
		\centering
		\includegraphics[width=1\linewidth]{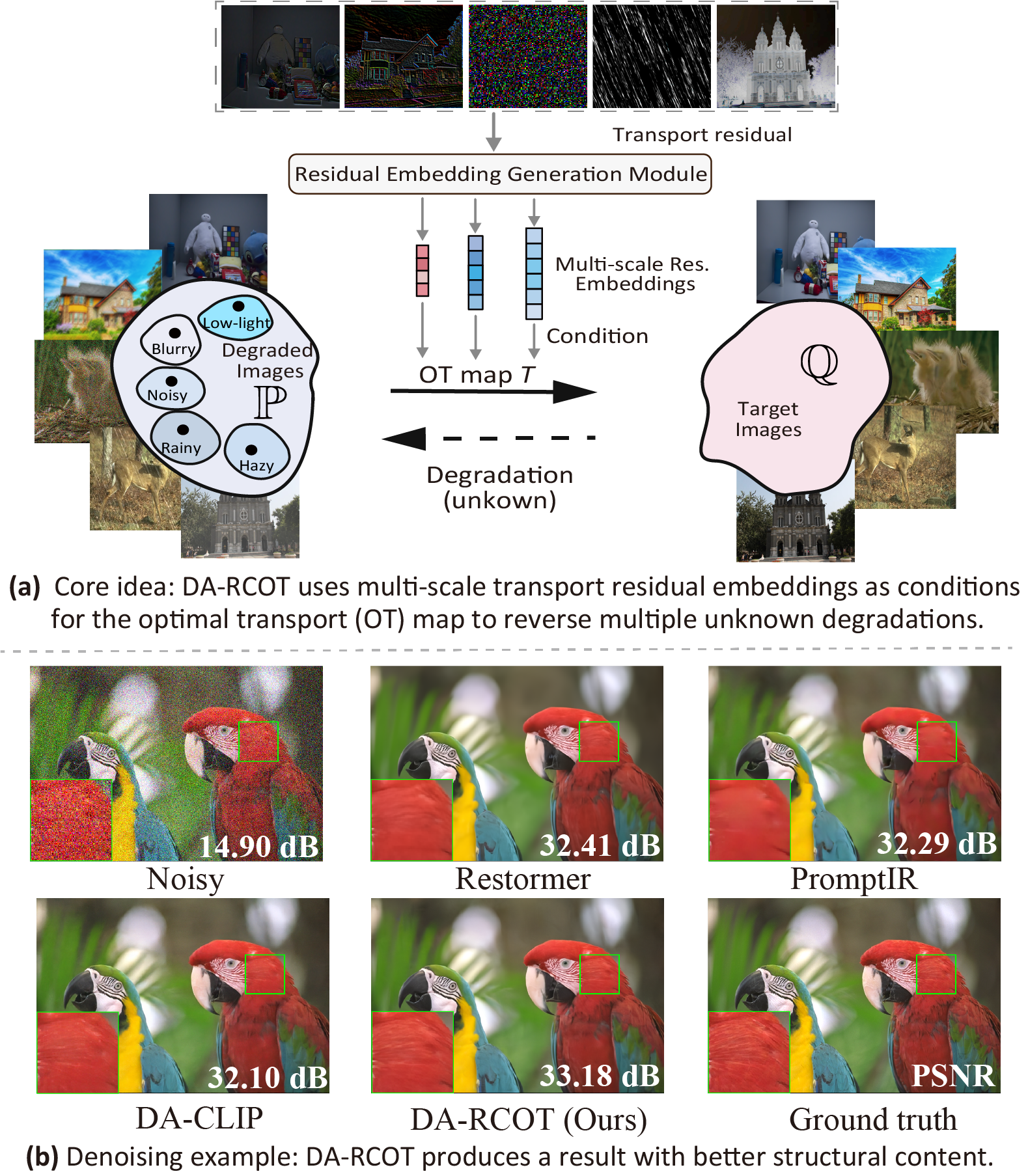}
		\vspace{-0.8cm}
		\caption{(a) The core idea of DA-RCOT is to model AIR as an OT problem and then condition the OT map with the customized multi-scale residual embeddings, yielding a degradation-aware and structure-preserving DA-RCOT map.  (b) A denoising demo under noise level $\sigma=50$. DA-RCOT produces a noise-free image with more faithful textures.}
		\label{demo}
		\vspace{-0.3cm}
	\end{figure}

	To tackle the AIR problem, several works \cite{fan2019general}, \cite{li2022all}, \cite{valanarasu2022transweather}, \cite{zhang2023ingredient}, \cite{potlapalli2023promptir}, \cite{luocontrolling}, \cite{zamfir2024efficient} directly fit joint regression models over multi-domain degraded-clean image pairs, minimizing pixel-wise distortion measures such as $L_1$ or $ L_2$ distances. Particularly, these methods derive degradation embeddings from the degraded image as task indicators, enabling the model to perform appropriate restoration while perceiving the degradation type. However, the severe ill-posedness of the regression with heterogeneous degradations still poses unique challenges for these methods, including perceiving the degradation (type and level) inadequately and leading to  "average" or suboptimal restoration maps. Consequently, they tend to produce results with compromised structural details \cite{blau2018perception}, \cite{ohayon2023reasons}, \cite{adrai2023deep}, e.g., distorted color and excessively smoothed textures (see Figure \ref{demo} (b)). Moreover, existing AIR methods rely on large amounts of paired data for supervised learning, which is usually conducted using synthetic data due to the difficulty in acquiring real-world degraded-clean pairs. As a result, the learned models tend to over-fit to the synthetic data and are vulnerable in real-world scenarios, where the availability of real-world image pairs, coupled with the complexity and diversity of real-world degradations, further hinders their applicability. 
	
	The challenges of AIR boil down to: \textit{i}) seeking the optimal transformation of heterogeneous degraded images into clean ones while \textit{ii}) properly perceiving the degradation semantics and prompting the preservation of visual structures, and \textit{iii}) extending to unpaired or partially paired data. 
	
	To address these challenges, this work is motivated to tackle the AIR problem from an optimal transport (OT) perspective for both unpaired and paired settings. The core idea is to seek the OT map that minimizes the transport distance between the distribution of multi-domain degraded images and the distribution of clean images, while informing the OT map with the novel transport residual (i.e., the domain gap between degraded and clean images) that contains degradation-specific knowledge about the degradation and image structures (see Figure \ref{demo} (a)). In this sense, the computed OT map has the potential to perceive the degradation (type and level) and further preserve the structural contents. 
	
	Specifically, we present a Degradation-Aware Residual-Conditioned Optimal Transport (DA-RCOT) approach, introducing the novel transport residual as a degradation-specific cue for both the transport cost and the transport map. Firstly, we formalize AIR as an OT problem between the distribution of multi-domain degraded images and the distribution of clean images with a Fourier residual-guided OT (FROT) objective that exploits the frequency patterns of the residual for the transport cost. Secondly and most crucially, we design the transport map as a two-pass DA-RCOT map, in which the transport residual is computed in the first pass and then encoded as multi-scale residual embeddings to condition the second-pass restoration.  This multi-scale conditioning mechanism injects intrinsic degradation knowledge (e.g., degradation type and level) and structural information from the multi-scale residual embeddings into the OT map, which thereby can dynamically adjust its behaviors for structure-preserving all-in-one restoration.
	
	Our contributions can be mainly summarized as follows:
	\begin{itemize}
		\item We advocate an OT solution dubbed DA-RCOT for AIR, learning to transform the distribution of multi-domain degraded images to that of clean ones with a minimal cost. Uniquely, DA-RCOT introduces the transport residual as a degradation-specific cue for both the transport cost and transport map.
		
		\item We integrate the degradation knowledge into the transport cost by exploiting the frequency patterns of transport residuals, yielding the FROT objective.
		
		\item We design the transport map as a two-pass DA-RCOT map, which dynamically conditions the OT map with degradation-specific low-level and high-level features from the multi-scale residual embeddings, enhancing its capability to perceive the degradation semantics and preserve the structures.
		
		\item Extensive experiments on synthetic and real-world data across five degradations show that DA-RCOT achieves state-of-the-art performance in all-in-one and task-specific settings. DA-RCOT also exhibits superior generalization to unseen degradation levels and robustness to the number of degradation types.
	\end{itemize}
	
	This work extends our conference paper \cite {tang2024residualconditioned} published at ICML. In the previous version, we learned a separate model based on OT theories to address single degradation, using single-scale residual embedding for structure preservation. In this work, we address the more challenging All-in-One problem by learning a unified OT map to handle different types of degradations simultaneously. Moreover, we extract multi-scale residual embeddings to condition the OT map for degradation-aware restoration. Specifically, we additionally make the following contributions: 
	1) We formulate AIR as an OT problem that seeks the optimal transportation from the distribution of multi-domain degraded images to that of the clean ones. 2) We employ gating-based Transformer blocks with a contrastive loss to extract multi-scale residual embeddings as conditions, which enhance the OT map with abundant degradation semantics and structural information.  3) We perform extensive experiments under AIR setting on a mixed dataset collected from multiple benchmarks, including Rain100L \cite{yang2017deep} for deraining, BSD68 \cite{martin2001database} for denoising, SOTS \cite{li2018benchmarking} for dehazing, GoPro \cite{nah2017deep} for deblurring, and LOL \cite{wei2018deep} for low-light enhancement. The results substantiate the effectiveness of the proposed method and demonstrate its robustness against varying degradation numbers and unseen degradation levels as compared to the conference version and other AIR methods.
	
	
	\vspace{-0.3cm}
	\section{Related Works}
	
	\subsection{Image Restoration}
	
	\textbf{Task-specific Image Restoration.} Task-specific IR focuses on learning separate models over degraded-clean image pairs to handle single degradation. A primary body of works \cite{Zamir2021MPRNet}, \cite{Zamir2021Restormer}, \cite{liang2021swinir}, \cite{chen2022simple}, \cite{luo2023image}, \cite{zhou2023fourmer}, \cite{ren2019progressive}, \cite{wang2023promptrestorer}, \cite{chen2023learning}, \cite{zhang2018learning}  are driven by efficient network architectures, optimizing the model under pixel-wise $L_1$ or $L_2$ distances to produce deterministic results, which often yields decent quantitative performance. 
	Another line of task-specific methods tackles image restoration from a distribution-fitting perspective with deep generative models \cite{li2018single}, \cite{pan2020physics}, \cite{chrysos2020rocgan}, \cite{mirza2014conditional}, \cite{bora2018ambientgan} and utilizes the degraded input as a condition for generation. RoCGAN \cite{chrysos2020rocgan} adapts CGAN \cite{mirza2014conditional} and uses high-quality data for supervised training of the restoration model. AmbientGAN \cite{bora2018ambientgan} generates clean images from noisy input, assuming the degradation satisfies certain constraints. Recently, diffusion models have emerged as a promising approach for generation. SR3 \cite{saharia2022image} conditions diffusion models on degraded images during training. Snips \cite{kawar2021snips}, DDRM \cite{kawar2022denoising}, and DPS \cite{chung2023diffusion} assume the degradation and its parameters are known at test time. IDM \cite{gao2023implicit} introduces a scale-adaptive condition on low-resolution images to achieve high-fidelity super-resolution, while IR-SDE \cite{luo2023image} proposes a loss function based on maximum likelihood to train a mean-reverting score-based model.  These methods, which use the degraded image as a condition for posterior distribution sampling, generally produce photo-realistic results. However, they rarely seek the optimal transformation between distributions and often disregard extra degradation knowledge for guidance. Moreover, they rely on large amounts of paired data for training regression models, and thus are prone to "average" results with compromised image structures.
	
	\textbf{All-in-One Image Restoration.} As compared to task-specific IR, AIR exhibits more potential for real-world applications but poses a unique challenge of addressing multiple types of degradations simultaneously. Current methods focus on how to extract informative degradation semantics from degraded input to promote task identification. AirNet \cite{li2022all} trains an extra encoder using contrastive learning to extract degradation embeddings from degraded images, which are then used to guide the restoration. IDR \cite{zhang2023ingredient} breaks down image degradations into their physical principles and implements AIR  using a two-stage meta-learning based approach. PromptIR \cite{potlapalli2023promptir} employs a learnable visual prompt module to adaptively encode the information of degradation type and further guide the restoration. DA-CLIP \cite{luocontrolling} uses the text embeddings from the fixed CLIP text encoder to adapt corresponding content embeddings and prompt-based degradation embeddings, which are subsequently integrated into large-scale restoration frameworks for AIR. These methods fit $L_1$- or $L_2$-based regression models and derive degradation embeddings from the degraded images, which often result in an inadequate perception of the degradation and lead to suboptimal restoration maps.
	
	On the contrary, our DA-RCOT seeks the OT map to transform the multi-domain degraded images into clean ones. Meanwhile, we introduce the novel transport residual as a degradation-specific cue for both the transport cost and transport map, promoting the OT map to adaptively perceive the degradation semantics and structural information.
	\vspace{-0.6cm}
	\subsection{Optimal Transport in Generative Models} 
	OT problem seeks to determine the optimal \textit{transport map} (Monge Problem \cite{monge1781memoire}, also known as MP) or \textit{transport plan} (Kantorovich Problem \cite{kantorovich1942translocation}, also known as KP) to transform a distribution into another with the minimal cost. 
	
	Recently, OT has been presented as a powerful toolkit or framework for machine learning, especially in the field of generative models. A majority of generative models (e.g., WGAN \cite{arjovsky2017wasserstein}, WGAN-GP \cite{gulrajani2017improved}) use OT cost as the loss for the generative network. A different set of approaches computes OT maps or plans for generative modeling. OTCS \cite{Gu2023optimal} computes optimal transport plans to guide the training of the conditional score-based diffusion model for super-resolution and translation. NOT \cite{korotin2023neural}, KNOT \cite{korotin2023kernel} estimate OT maps and plans using neural networks under the duality framework and apply their method to unpaired image translation. OTUR \cite{wang2022optimal} relaxes the transport constraint in Monge formulation with a Wasserstein-1 discrepancy between the target distribution and the transported distribution from noisy input. Despite pioneering the way of modeling domain translation as an OT problem, these methods rarely incorporate the degradation knowledge and empirically use a pixel-wise $L_2$  transport cost which may not serve as an efficient metric over complex degraded/clean image manifolds. In contrast, our DA-RCOT integrates the degradation-specific transport residual and its multi-scale embeddings into the OT framework, resulting in the FROT objective and degradation-aware DA-RCOT map for AIR.
	
	\vspace{-0.3cm}
	\section{Preliminaries}
	\textbf{Notation.} Let $Y$ and $X$ be Polish spaces, and let $\mathcal{P}(Y)$ and $\mathcal{P}(X)$ denote the corresponding sets of probability measures on these spaces. We define $\Pi(\mathbb{P}, \mathbb{Q})$ as the set of all joint probability distributions on $Y \times X$ with marginals $\mathbb{P}$ and $\mathbb{Q}$. For any measurable map $T: Y \rightarrow X$, the push-forward operator associated with $T$ is denoted as $T_\#$. We use $\|\cdot\|$ to represent the standard Euclidean norm unless specified otherwise. Lastly, the space of functions on $Y$ that are integrable with respect to $\mathbb{Q}$ is denoted by $L^1(\mathbb{Q})$.
	
	Given two distributions $\mathbb P\in \mathcal P(X)$ and $\mathbb Q\in\mathcal P(Y)$, along with a transport cost $c: Y \times X \rightarrow \mathbb{R}_+$, the OT problem seeks the least-effort transport from $\mathbb P$ to $\mathbb Q$.
	
	\textbf{Definition 1 (Transport map)}: $T : Y \to X$ is a transport map that transports $\mathbb P$ to $\mathbb Q$ if
	\[
	\mathbb P(A) = \mathbb Q(T^{-1}(A)),
	\]
	for all $\mathbb P$-measurable sets $A$.
	
	\textbf{Definition 2 (MP)}: the Monge problem \cite{monge1781memoire} for optimal transport can be formulated as follows:
	\begin{align}
		\label{monge}
		\text{MP-Cost}(\mathbb{P},\mathbb{Q})\triangleq\inf_{T_\#\mathbb{P}= \mathbb{Q}}\int_{Y}c\big(y,T(y)\big)d\mathbb{P}(y),
	\end{align}
	where \( T: Y \to X \) is a transport map that pushes forward \( \mathbb{P} \) to \( \mathbb{Q} \), denoted as \( T_{\#} \mathbb{P} = \mathbb{Q} \). The map $T^*$ attaining the infimum is called the \textit{optimal transport map}, which is taken over all the transport maps. Since the map $T$ satisfying $T_\#\mathbb P=\mathbb Q$ sometimes does not exist, Kantorovich proposes the convex relaxation of the above Monge problem.
	
	\textbf{Definition 3 (KP)}: the Kantorovich problem \cite{kantorovich1942translocation} for optimal transport can be formulated as follows:
	\begin{align}
		\text{KP-Cost}(\mathbb{P},\mathbb{Q})\triangleq\inf_{\pi\in\Pi(\mathbb P,\mathbb Q)}\int_{Y\times X}c(y,x)d\pi(y,x),
		\label{Kon}
	\end{align}
	where $\pi\in\Pi(\mathbb P,\mathbb Q)$ is a transport plan, i.e., the joint distribution on $Y\times X$ whose marginals are $\mathbb P$ and $\mathbb Q$. The plan $\pi^*$ attaining the infimum is called the \textit{optimal transport plan}, which is taken over all the transport plans.
	
	\textbf{Definition 4 (DP)}: KP has the dual form \cite{villani2009optimal}:
	\begin{align}
		\text{DP-Cost}(\mathbb{P},\mathbb{Q})=\sup_{\varphi}\int_Y\varphi^{c}(y)d\mathbb P(y)+\int_X\varphi(x)d\mathbb Q(x), 
		\label{dual}
	\end{align}
	where $\displaystyle\varphi^{c}(y)=\inf_{x\in X}\left[c(x,y)-\varphi(x)\right]$ is the $c$-transform of $\varphi$ and the optimal potential $\varphi^*$ is taken over all $\varphi\in L^1(\mathbb Q)$.
	\vspace{-0.2cm}
	\section{Method}
	\begin{figure*}[!t]
		\centering
		\includegraphics[scale=2.0]{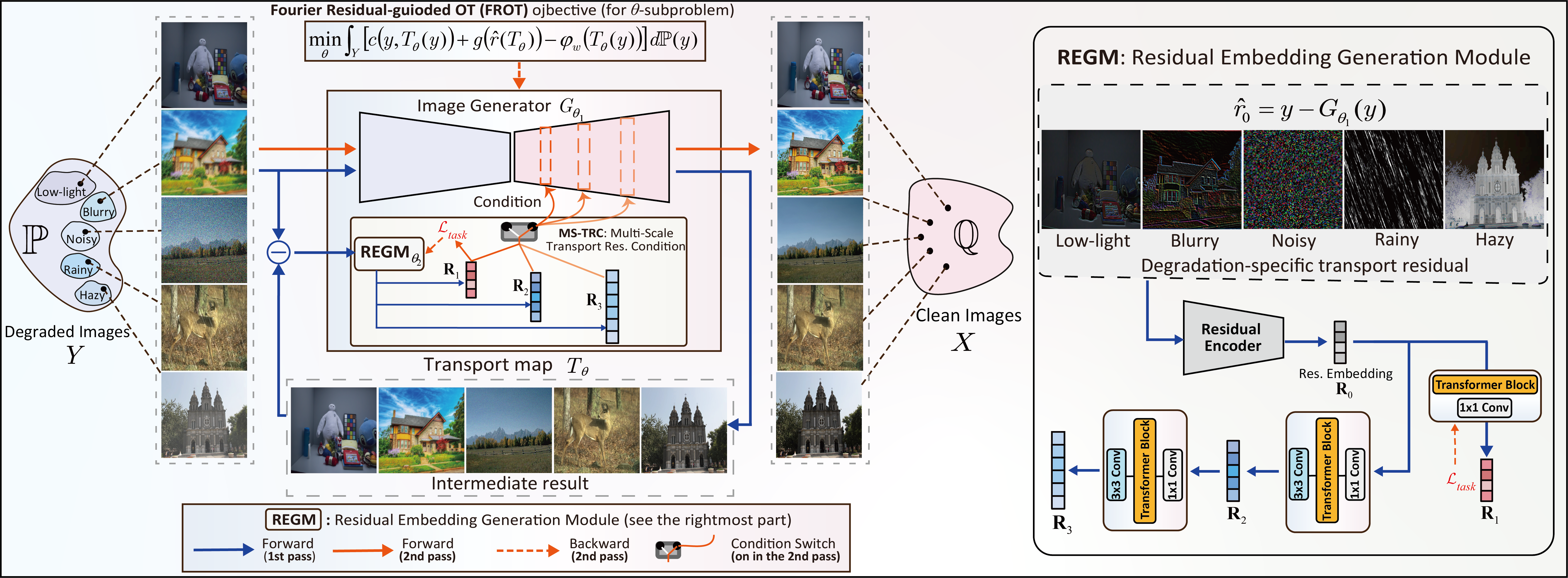}
		\vspace{-0.3cm}
		\caption{Overview of the DA-RCOT framework for AIR.  DA-RCOT integrates the transport residual into the transport cost, yielding the FROT objective; and more crucially, into the transport map via a two-pass conditioning process. The first pass unconditionally generates an intermediate result along with the residual $\hat r_0$. Then the residual is encoded and adapted as multi-scale embeddings $ \{\mathbf R_i\}_{i=1,2,3}$ to condition the second-pass restoration. }
		\label{model}
		\vspace{-0.5cm}
	\end{figure*}
	In this section, we present DA-RCOT, an OT framework to learn a structure-preserving and degradation-aware OT map for AIR. The key idea is to model AIR as an OT problem and introduce the transport residual as a degradation-specific cue for both the transport cost and transport map. 
	
	\textbf{Method overview.} We first model AIR as an OT problem, exploiting the frequency patterns of the residual, yielding the FROT objective that measures the distance between the distribution of multi-domain degraded images and the clean one (section 4.1). Secondly, we integrate the degradation-specific knowledge from the multi-scale residual embeddings into the transport map via a two-pass conditioning process (section 4.2), in which the transport residual is computed in the first pass and then encoded as multi-scale residual embeddings to condition the second-pass restoration (see Figure \ref{model}). Section 4.3 illustrates the learning of multi-scale residual embeddings, while section 4.4 presents the t-SNE visualization of the dense residual embeddings for task identification. In section 4.5, we summarize the algorithm for computing the DA-RCOT map by adversarially training two neural networks to solve the minimax problem. 
	\vspace{-0.2cm}	
	\subsection{Residual-guided OT formulation for AIR}
\begin{figure*}[!t]
		\centering
		\includegraphics[scale=0.65]{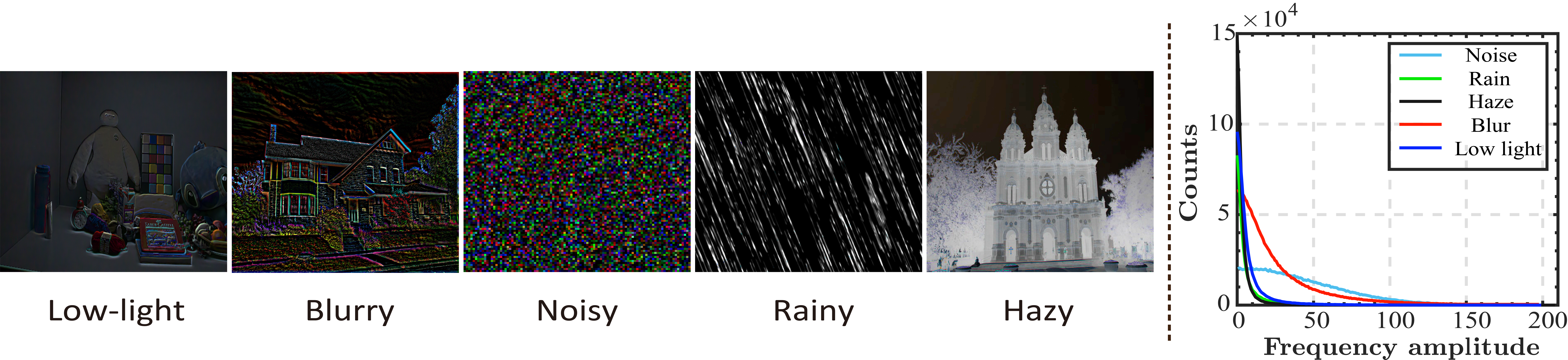}
		\vspace{-0.3cm}
		\caption{Left: Visual examples of the transport residual $r$. Right: Counts of the frequency amplitude of residuals for five types of degradation. For all degradations except for the noise, the residuals are generally sparse in the Fourier domain. The curves are averaged with 40 degraded-clean pairs. }
		\label{anl}
		\vspace{-0.5cm}
	\end{figure*}

	Let $\mathbb P$ and $\mathbb Q$ represent the distributions of the degraded image domain $Y$ and clean image domain $X$, respectively. In the context of AIR, the degraded image domain $Y$ contains multi-domain images of different degradations. To incorporate the degradation knowledge in the OT framework, we introduce a residual regularization term $g(\cdot)$ for the transport cost $c(y,x)$, characterizing the frequency patterns of the transport residual $r=y-x$, i.e., the degradation domain gap. Finally, we model AIR based on KP (\ref{Kon}) with the residual regularizer in the transport cost, yielding the Fourier residual-guided OT, i.e., FROT objective:
	\begin{align}
		\label{FROT}
		\text{KP-FROT}(\mathbb P, \mathbb Q)  \triangleq \inf_{\pi\in {\rm \Pi}(\mathbb P,\mathbb Q)}\int_{Y\times X}\tilde{c}(y,x)d\pi(y,x).
	\end{align}
	Here $\tilde{c}(y,x)=c(y,x)+g(r)$, and $c(y,x)$ is chosen to be the standard Euclidean distance, while $g(\cdot):Y\times X\rightarrow \mathbb R_+$ is determined according to the frequency patterns shown in Figure \ref{anl}. The transport residual $r$ exhibits degradation-specific patterns in the image domain. While in the frequency domain, the transport residuals for degradations, except for noise, generally show sparsity. In light of this, we use $\ell_1$ norm to penalize the Fourier residual for all the degradations except for noise, i.e., $g(r)=\|\mathcal F(r)\|_1$, and particularly use $\ell_2$ norm for the denoising task.
	
	With Eq. (\ref{FROT}), we intend to find the OT map $T^*$ from $Y$ to $X$, which can flexibly map degraded images into clean ones. Since the duality (\ref{dual}) can lead us to a manageable approach, we derive the following dual form from Eq. (\ref{FROT}) 
	\begin{align}
		\nonumber\text{DP-FROT}(\mathbb P, \mathbb Q)=\sup_{\varphi}\int_Y\varphi^{\tilde{c}}(y)d\mathbb P(y)+\int_X\varphi(x)d\mathbb Q(x), 
	\end{align}
	where $\displaystyle\varphi^{\tilde{c}}(y)=\inf_{x\in X}\left[c(x,y)+ g(r)-\varphi(x)\right]$.
	Replacing the optimization of the $\varphi^{\tilde{c}}(y)$ term over target $x\in X$ with an equivalent optimization (Rockafellar interchange theorem \cite{rockafellar1976integral}, Theorem 3A) over the map of interest $T: Y\rightarrow X$,  we obtain the minimax reformulation of dual form:
	\begin{align}
		\label{minimax}
		\nonumber&\text{DP-FROT}(\mathbb P, \mathbb Q)=\sup_{\varphi}\inf_{T}\bigg\{\mathcal L_{u}(T,\varphi)\triangleq\int_X\varphi(x)d\mathbb Q(x)\\
		&+\int_{Y}\big[-\varphi(T(y))+\underbrace{c(y,T(y))+ g(\hat r(T))}_{\text{transport cost}:~ \tilde c(y,T(y))}\big]d\mathbb P(y)\bigg\},
	\end{align}
	where $\hat r(T)=y-T(y)$ estimates the transport degradation domain gap (termed as transport residual). Notably, only unpaired data is required for solving the optimization problem (\ref{minimax}). To further leverage the information of paired data, which is commonly encountered in practical applications, we introduce the following pair-informed FROT objective.
	
	\textbf{Pair-informed FROT Formulation.} Given a paired data subset of $Y\times X$, consisting of $n$ degraded-clean image pairs $(y_1,x^*(y_1)),\dots,(y_n,x^*(y_n))$, we can rewrite Eq. (\ref{FROT}) with  the pairwise constraint into the transport cost:
	\begin{align}
		\label{pair-FROT}
		\inf_{\pi\in {\rm \Pi}(\mathbb P,\mathbb Q)}\int_{Y\times X}\left[\tilde{c}(y,x)+\lambda\ell(x,x^*(y))\right]d\pi(y,x).
	\end{align}
	Similarly to the previous steps, from Eq. (\ref{pair-FROT}) we can derive the minimax problem under such pairwise constraints, and the corresponding paired FROT objective function is
	\begin{align}
		\label{pair-func}
		\nonumber \mathcal L_{p}(T,\varphi)\triangleq \int_X\varphi(x)d\mathbb Q(x)+\int_{Y}\big[
		-\varphi(T(y))\\+\underbrace{c(y,T(y))+g(\hat r(T))+\lambda\ell(T(y),x^*(y))}_{\text{transport cost}:~\tilde C_{x^*}(y,T(y))}\big]d\mathbb P(y).
	\end{align}
	The function $\ell:Y\times X\rightarrow \mathbb R_+$ is a loss metric measuring the distance between samples. In our experiments, we employ the $\ell_1$ norm, i.e., $\ell(x,\hat x)=\|x-\hat x\|_1$.	For simplicity, we denote the transport cost terms in (\ref{minimax}) and (\ref{pair-func}) with $\tilde c(y,T(y))$ and $\tilde C_{x^*}(y,T(y))$, then the unpaired and paired FROT minimax problems can be simplified as:
	\begin{align}
		\label{unpair-mini}
		\nonumber&\text{DP-FROT}(\mathbb P, \mathbb Q)=\sup_{\varphi}\inf_{T}\bigg\{\mathcal L_{u}(T,\varphi)\triangleq\int_X\varphi(x)d\mathbb Q(x)\\
		&+\int_{Y}\left[\tilde c(y,T(y))-\varphi(T(y))\right]d\mathbb P(y)\bigg\}~~ (\text{Unpaired}),
		\\
		\label{pair-mini}
		\nonumber&\text{DP-FROT}(\mathbb P, \mathbb Q)=\sup_{\varphi}\inf_{T}\bigg\{\mathcal L_{p}(T,\varphi)\triangleq\int_X\varphi(x)d\mathbb Q(x)\\
		&+\int_{Y}\left[\tilde C_{x^*}(y,T(y))-\varphi(T(y))\right]d\mathbb P(y)\bigg\}~~ (\text{Paired}),
	\end{align}
	We can show that solving the FROT minimax problems provides OT maps, for both unpaired and paired cases. 
	\noindent\textbf{Proposition 1.}	(Saddle points of FROT provide OT maps). For any optimal potential $\varphi_1^*\in\arg\sup_\varphi\mathcal L_u(T,\varphi_1)$ and $\varphi_2^*\in\arg\sup_\varphi\mathcal L_p(T,\varphi_2)$, it holds for the  OT map $T^*$ that
	\begin{align}
		\label{saddle} T^*\in \mathop{\arg\inf}_{T:Y\rightarrow X}\mathcal L_u(T,\varphi_1^*), \\
		\label{pair-saddle} T^*\in \mathop{\arg\inf}_{T:Y\rightarrow X}
		\mathcal L_p(T,\varphi_2^*).
	\end{align}
	\textit{Proof}. We provide proof for the unpaired case (\ref{saddle}), and the proof for the paired case (\ref{pair-saddle}) is identical, so it is omitted. According to the Monge problem (\ref{monge}), the MP-FROT objective under transport cost $\tilde c$ is
	\begin{align} 
		\label{Monge-fort}
		\text{MP-FROT}(\mathbb P, \mathbb Q)=\inf_{T_{\#}\mathbb P=\mathbb Q}\int_Y\tilde c(y,T(y))d\mathbb P(y). \end{align}
	Since $\varphi_1^*$ is the optimal potential, we have $\inf_{T}\mathcal L(T,\varphi_1^*)=\text{FROT}(\mathbb P, \mathbb Q)$.
	For the OT map $T^*$ in (\ref{Monge-fort}), using $T^*_{\#}\mathbb P=\mathbb Q$ and the change of variables $T^*(y)=x$, we deduce
	$$\int_Y\varphi_1^*(T^*(y))d\mathbb P(y)=\int_X\varphi_1^*(x)d\mathbb Q(x).$$
	Substituting this equation into $\mathcal L_u(T,\varphi_1)$, we obtain
	\begin{align}\label{minLu}\mathcal L_u(T^*,\varphi_1^*)&=\int_Y\tilde c(y,T^*(y))d\mathbb P(y)\\&=\inf_{T_{\#}\mathbb P=\mathbb Q}\int_Y\tilde c(y,T(y))d\mathbb P(y)=\text{MP-FROT}(\mathbb P, \mathbb Q).\nonumber\end{align}
	According to Pratelli \cite{PRATELLI20071} and Villani \cite{villani2009optimal}, since $\tilde c$ is continuous and $(X,\mathbb P)$ is Polish without atoms, we have $$\text{MP-FROT}(\mathbb P,\mathbb Q)=\text{KP-FROT}(\mathbb P,\mathbb Q)=\text{DP-FROT}(\mathbb P,\mathbb Q).$$ This result, together with Eq. (\ref{minLu}), demonstrates that $T^*\in\mathop{\arg\inf}_{T}\mathcal L_u(T,\varphi_1^*)$. For the paired case with $\mathcal L_p$ and $\varphi_2^*$, we immediately reach the conclusion by replacing $\tilde c$ with $\tilde  C_{x^*}$, which is also continuous, thus completing the proof. \qed
	
	Proposition 1 affirms the feasibility of solving the minimax problems to acquire an optimal pair, constituting an OT map from $\mathbb P$ to $\mathbb Q$.  For general $\mathbb P$ and 
	$\mathbb Q$, given some optimal potential $\varphi^*$, the set $\arg\inf_T$  may encompass not only the OT map $T^*$ but also other saddle points, which are capable of delivering decent performance as in experiments (section \ref{exp}). To tackle the minimax problems, we parameterize the transport map $T$ and potential $\varphi$ using two neural networks  $T_\theta$ and $\varphi_\omega$, and then perform adversarial training to optimize their parameters against each other.
	\subsection{Learning Degradation-Aware OT Map Conditioned on Multi-Scale Residual Embeddings}
	In order to enable the OT map to perceive the degradation semantics for AIR while preserving more image structures, we design the transport map as the two-pass DA-RCOT map by taking the multi-scale residual embeddings as its additional conditions. The transport residuals convey degradation-specific patterns related to both the degradation (type and level) and image structures (see Figure \ref{anl}), which could be leveraged to inform the OT map for degradation-aware and structure-preserving restoration. To this end, the parameterized transport map $T_\theta$ is implemented to comprise two components: an image generator $G$ with parameters $\theta_1$, and a multi-scale residual embeddings generation module (REGM) with parameters $\theta_2$.  We next elaborate the two-pass restoration process of the DA-RCOT map (see the overall framework shown in Figure \ref{model}).
	
	\noindent\textbf{Two-pass DA-RCOT map for restoration.} Given a degraded image $y$ from the heterogeneous degraded image domain $Y$, the first pass of the DA-RCOT map unconditionally generates an intermediate restored result $\hat x_0=G(y;\theta_1)$ via the generator and calculates the corresponding intermediate transport residual $\hat r_0=y-\hat x_0$. Then the residual $\hat r_0$ is fed into REGM to produce multi-scale residual embeddings $\{\mathbf R_i\}_{i=1,2,3}$. In the second pass, these embeddings serve as conditions for the generator $G$, yielding the refined result $T_\theta(y)=G(y|\mathbf R_1, \mathbf R_2, \mathbf R_3;\theta_1).$
	This process boils down to:
	\begin{align}
		\hat r_0=y-&G(y;\theta_1);\left[\mathbf{R}_1,\mathbf{R}_2,\mathbf{R}_3\right]=\texttt{REGM}(\hat r_0;\theta_2);\nonumber\\
		&T_\theta(y)=G\left(y|\mathbf{R}_1,\mathbf{R}_2,\mathbf{R}_3;\theta_1\right).\label{m2p}
	\end{align}
	where $\{\mathbf R_i\}_{i=1,2,3}$ serve as conditions for different scales of the generator $G$, with $\mathbf R_1$ specifically used at the dense scale to enhance the semantics of heterogeneous degradations.
	
	\noindent\textbf{Learning multi-scale residual embeddings with REGM.} To extract the multi-scale residual embeddings $\{\mathbf R_i\}_{i=1,2,3}$, the REGM incorporates some gating-based Transformer blocks that selectively permit the propagation of essential features across different scales. The blocks consist of two key sub-modules: Multi-Dconv head transposed attention (MDTA), and Gated-Dconv feedforward network (GDFN) \cite{Zamir2021Restormer}. MDTA uses depth-wise convolutions to emphasize the local image structures, while GDFN selectively transforms features with a gating mechanism that suppresses the less informative features and allows only useful ones to propagate. The propagation process of REGM (see the rightmost of Figure \ref{model}) is summarized as 
	\begin{align}
		\mathbf{R}_0=&\texttt{Encoder}(\hat{r}_0);~\mathbf{R}_1=\left(\texttt{GDFN}\left(\texttt{MDTA}(\texttt{Conv}_{1\times1}\left(\mathbf R_0 \right)\right)\right);\nonumber\\
		&\mathbf{R}_2=\texttt{Conv}_{3\times3}\left(\texttt{GDFN}\left(\texttt{MDTA}\left(\texttt{Conv}_{1\times1}\left(\mathbf R_{0}\right)\right)\right)\right);\nonumber\\
		&\mathbf{R}_3=\texttt{Conv}_{3\times3}\left(\texttt{GDFN}\left(\texttt{MDTA}\left(\texttt{Conv}_{1\times1}\left(\mathbf R_{2}\right)\right)\right)\right). 
		\label{emd}
	\end{align}
	Moreover,  the work \cite{voynov2023p+} observes that the shallow layers of U-shaped architecture capture low-level structural features such as color and textures, while the deep layers provide semantic guidance. In light of this, we employ a contrastive loss that promotes $\mathbf R_1$ to emphasize the task semantic features of the degradation. Specifically, for the residual embedding $\mathbf R_1^k$ from $k$-th task among $K$ tasks, the positive samples are embeddings from other degraded inputs of $k$-th task, denoted as $\mathbf R_1^{k+}$; and the negative samples are embeddings from other tasks $\mathbf R_1^{k-}$. The contrastive loss for task identification is defined as
	\begin{small}
		\begin{equation}
			\mathcal L_{task}=-\sum_{k=1}^{K}\log\frac{\displaystyle\sum_{~~\mathbf R_1^{k+}}\exp(\mathbf R_1^k\cdot \mathbf R_1^{k+}/\tau)}{\displaystyle\sum_{~~\mathbf R_1^{k+}}\exp(\frac{\mathbf R_1^k\cdot \mathbf R_1^{k+}}{\tau})+\sum_{~~\mathbf R_1^{k-}}\exp(\frac{\mathbf R_1^k\cdot \mathbf R_1^{k-}}{\tau})},
			\label{task}
		\end{equation}
	\end{small}
	
	\noindent where $\tau$ is the temperature hyper-parameter. As shown in Figure \ref{model}, the learned residual embedding $\mathbf R_1$ is utilized as the condition at the dense scale to enhance the OT map's capability to perceive the degradation semantics. Meanwhile, $\mathbf R_2$ and $\mathbf R_3$ are used at shallow layers to augment the OT map with extra low-level structural information. 
	\begin{figure}[!t]
		\centering
		\includegraphics[scale=0.48]{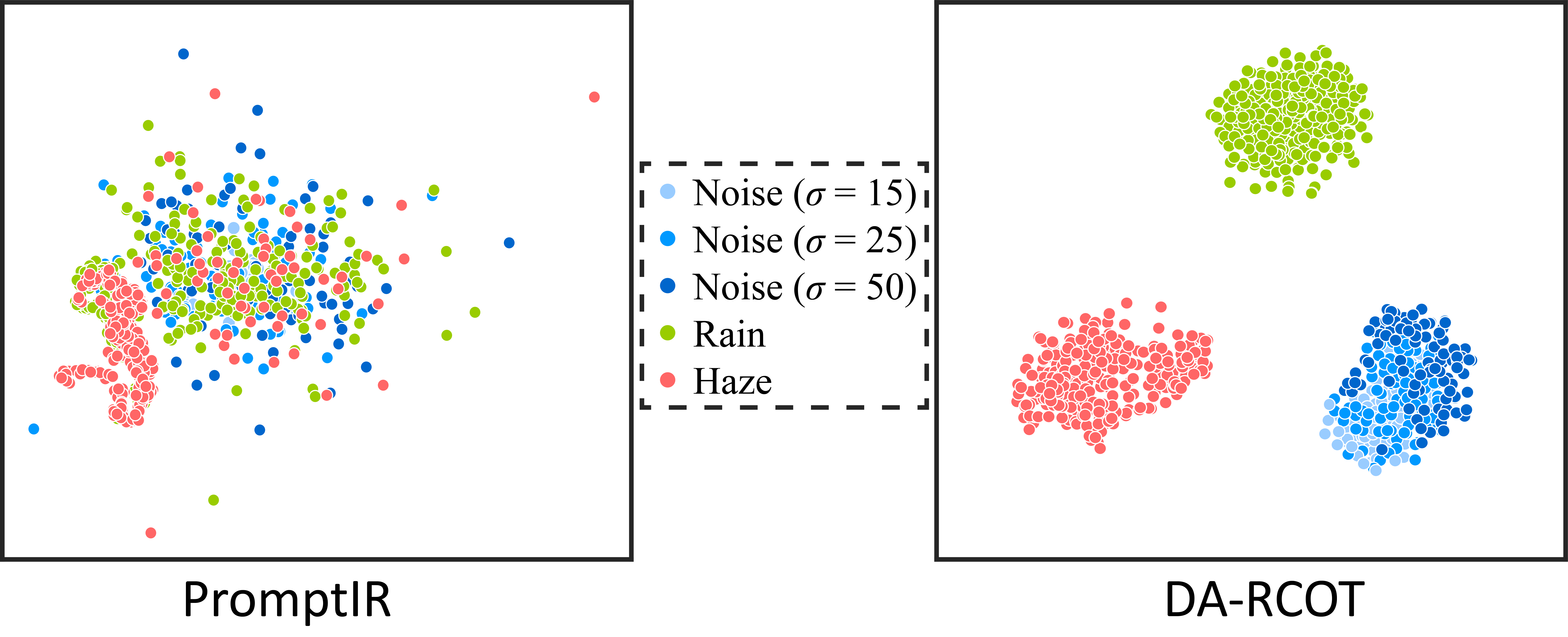}
		\vspace{-0.3cm}
		\caption{The t-SNE visual comparison of the degradation embeddings from PromptIR \cite{potlapalli2023promptir} and DA-RCOT. As compared to the prompt-based embeddings of PromptIR \cite{potlapalli2023promptir}, the residual embeddings $\mathbf R_1$ of DA-RCOT are clearly separated according to the specific tasks. Particularly, the residual embeddings \textit{w.r.t.} different levels of noise are clustered together but also exhibit level-specific positional relationships.  }
		\vspace{-0.3cm}
		\label{emd_vis}
	\end{figure}
	
	\subsection{Visualization on the Degradation Embeddings}
	To better understand the semantic nature of the residual embedding $\mathbf R_1$, we employ t-SNE for visualizing, using 300 noisy images (100 each for $\sigma=15, \sigma=25, \sigma=50$), 300 rainy images, and 300 haze images. Embeddings from different sources are marked with distinct colors.
	
	We compare our residual embeddings with the prompt-based degradation embeddings from the state-of-the-art method PromptIR \cite{potlapalli2023promptir}. As shown in Figure \ref{emd_vis}, compared to the prompt-based degradation embeddings in PromptIR \cite{potlapalli2023promptir}, the residual embeddings $\mathbf R_1$ of DA-RCOT are clearly separated according to specific tasks. Particularly, the residual embeddings \textit{w.r.t.} different levels of noise are clustered together but also exhibit level-specific positional relationships. This observation demonstrates that $\mathbf R_1$ not only captures relatively precise semantics of the degradation but also retains sensitivity to perceive varying degradation levels, which enhances its robustness to handle diverse degradations in the presence of level shifts.
	\subsection{Overall Training Algorithm} 
	After parameterizing $T_\theta$ and $\varphi_w$, the objective functions of minimax problems (\ref{saddle}) and (\ref{pair-saddle}) can be rewritten as
	\begin{align}
		\mathcal{L}_{u}(\omega,\theta)=\mathbb{E}_{x\sim \mathbb{Q}}\left[\varphi_\omega(x)\right]+ \mathbb{E}_{y\sim \mathbb{P}}\big[ \tilde c(y,T_\theta(y))-\varphi_\omega(T_\theta(y)) \big],\nonumber
		\\
		\mathcal{L}_{p}(\omega,\theta)=\mathbb{E}_{x\sim \mathbb{Q}}\left[\varphi_\omega(x)\right]+ \mathbb{E}_{y\sim \mathbb{P}}\big[ \tilde  C_{x^*}(y,T_\theta(y))-\varphi_\omega(T_\theta(y)) \big].
	\end{align}
	For these two problems, we train the networks $T_\theta$ and $\varphi_w$ by respectively minimizing and maximizing the objectives $\mathcal L_u$ or $\mathcal L_p$,  i.e., $\max_{\omega}\min_{\theta} \mathcal{L}_{u}(\omega,\theta)$ or $\max_{\omega}\min_{\theta} \mathcal{L}_{p}(\omega,\theta)$. This can be implemented by adversarially training $T_\theta$ and $\varphi_w$, in which we estimate the expectation using mini-batch data in each training step. The training algorithm for computing the DA-RCOT map is summarized in Algorithm \ref{algo}.
	\begin{algorithm}[!t]
		\caption{Computing the DA-RCOT map for AIR.}  
		\textbf{Input}:  degraded images $Y\sim \mathbb P$; clean images $X\sim \mathbb Q$; transport network: $T_\theta$; potential network: $\varphi_w$;\\ the number of iterations of $\theta$ per iteration of $\omega$: $n_T$;
		
		\begin{algorithmic}[1]
			\label{algo}
			\WHILE{$\theta$ has not converged}
			\STATE Sample batches $\mathcal Y$ from $Y$, $\mathcal X$ from $X$;
			\STATE $\mathcal L_\varphi\leftarrow \frac{1}{|\mathcal Y|}\sum_{y\in \mathcal Y}\varphi_w(T_\theta(y))-\frac{1}{|\mathcal X|}\sum_{x\in \mathcal X}\varphi_w(x)$;
			\STATE	Update $w$ by using $\frac{\partial \mathcal L_\varphi}{\partial w}$;
			\FOR{$t=0,\cdots,n_T$}
			\STATE Compute $\{\mathbf R_i\}_{i=1,2,3}$ via (\ref{emd}); 
			\STATE Compute $\mathcal L_{task}$ via (\ref{task});
			\STATE Compute $T_\theta(y)$ via (\ref{m2p});
			\IF{there exist paired samples}
			\STATE $\mathcal L_T\leftarrow\frac{1}{|\mathcal Y|}\sum_{\substack{y \in \mathcal{Y} \\[-0.1em] \scriptsize{x^* \in \mathcal{X}}}}\big[ \tilde  C_{x^*}(y,T_\theta(y))-\varphi_\omega(T_\theta(y)) \big];$
			\ELSE
			\STATE $\mathcal L_T\leftarrow\frac{1}{|\mathcal Y|}\sum_{y\in \mathcal Y}\big[ \tilde c(y,T_\theta(y))-\varphi_\omega(T_\theta(y)) \big];$
			\ENDIF
			\STATE $\mathcal L\leftarrow\mathcal L_T + \gamma \mathcal L_{task}$;
			\STATE Update $\theta$ by using $\frac{\partial \mathcal L}{\partial \theta}$;
			\ENDFOR
			\ENDWHILE
		\end{algorithmic} 
	\end{algorithm} 
	
	\section{Experiments}
	\begin{table*}[!h]
		\centering
		\caption{The \textit{All-in-One} comparison of our DA-RCOT with the state-of-the-art methods on \textbf{three} degradations. The metrics are reported in the form of PSNR($\uparrow$)/SSIM($\uparrow$)/LPIPS($\downarrow$)/FID($\downarrow$). ($*$) indicates the method is performed in an unpaired setting.}
		\label{air3}
		\vspace{-0.2cm}
		\setlength{\tabcolsep}{3pt}
		\renewcommand{\arraystretch}{1.2}
		\resizebox{\textwidth}{!}{
			\begin{tabular}{@{}clcccccc@{}}
				\toprule
				&\multirow{2}{*}{Method} 
				&\textit{Dehazing} & \textit{Deraining} & \multicolumn{3}{c}{\textit{Denoising}} 
				& \multirow{2}{*}{Average}  \\  \cmidrule(lr){3-3} \cmidrule(lr){4-4} \cmidrule(lr){5-7} 
				&& SOTS &Rain100L & BSD68\textsubscript{$\sigma$=15} & BSD68\textsubscript{$\sigma$=25} & BSD68\textsubscript{$\sigma$=50} &  \\
				\midrule
				\multirow{5}{*}{Unpaired}&CycleGAN$^*$\cite{zhu2017unpaired}    & 24.26/0.923/0.066/48.33 &  30.12/0.912/0.087/65.24  &  30.25/0.872/0.101/89.23&  26.75/0.776/0.174/112.6&   23.45/0.702/0.234/152.5 & 26.97/0.837/0.132/93.57  \\
				&WGAN-GP$^*$ \cite{gulrajani2017improved} & 25.28/0.929/0.043/34.98&32.87/0.970/0.066/51.12&30.87/0.869/0.104/71.32&28.12/0.801/0.158/97.22&26.18/0.722/0.216/130.2&28.66/0.857/0.117/76.97\\
				&OTUR$^*$ \cite{wang2022optimal}   & 25.56/0.935/0.046/39.59 & 32.83/0.934/0.076/58.91 & 31.17/0.889/0.099/76.13& 28.66/0.820/0.153/92.54& 26.53/0.744/0.211/126.7 & 28.95/0.864/0.117/78.76  \\
				\cmidrule{2-8}
				& RCOT$^*$ \cite{tang2024residualconditioned}  & \underline{27.02}/\underline{0.945}/\underline{0.038}/\underline{32.26}&\underline{33.56}/\underline{0.945}/\underline{0.055}/\underline{37.25}&\underline{31.95}/\underline{0.902}/\underline{0.085}/\underline{61.29}&\underline{29.33}/\underline{0.834}/\underline{0.135}/\underline{82.15}&\underline{26.98}/\underline{0.757}/\underline{0.201}/\underline{113.2}&\underline{29.77}/\underline{0.877}/\underline{0.103}/\underline{65.23}\\
				&DA-RCOT$^*$   & \textbf{27.64}/\textbf{0.964}/\textbf{0.027}/\textbf{23.17} & \textbf{33.82}/\textbf{0.960}/\textbf{0.032}/\textbf{28.54} & \textbf{32.28}/\textbf{0.918}/\textbf{0.064}/\textbf{40.58} & \textbf{30.12}/\textbf{0.864}/\textbf{0.122}/\textbf{72.21} & \textbf{27.16}/\textbf{0.779}/\textbf{0.189}/\textbf{101.7} & \textbf{30.20}/\textbf{0.897}/\textbf{0.087}/\textbf{53.24}
				\\
				\midrule
				\multirow{10}{*}{Paired}&MPRNet \cite{Zamir2021MPRNet}    & 25.43/0.956/0.038/28.15 &  33.66/0.955/0.057/33.65  &  33.50/0.925/0.084/52.87&  30.89/0.880/0.127/79.53&   27.48/0.778/0.201/121.9 & 30.19/0.899/0.101/63.23  \\
				&Restormer \cite{Zamir2021Restormer}   & 29.92/0.970/0.035/22.29 &  35.64/0.971/0.036/33.97  &  33.81/0.932/0.078/42.61&  31.00/0.880/0.113/74.62&   27.85/0.792/0.198/117.6 & 31.62/0.909/0.092/58.22 \\
				&IR-SDE \cite{luo2023image}   & 29.35/0.961/0.029/19.80 &  34.87/0.958/0.031/30.36  &  32.89/0.903/0.068/35.51	&  30.56/0.861/0.107/68.15&   27.22/0.769/0.195/107.6 & 30.98   0.890/0.086/52.29\\
				\cmidrule{2-8}
				&DL \cite{fan2019general} & 26.92/0.931/0.042/30.05&32.62/0.931/0.058/39.29 & 33.05/0.914/0.079/56.13&30.41/0.861/0.129/81.51 & 26.90/0.740/0.204/123.2 &29.98/0.875/0.102/66.04\\
				&AirNet \cite{li2022all} & 27.94/0.962/0.030/26.71&34.80/0.962/0.054/33.28&33.92/\underline{0.933}/0.066/45.67&31.26/0.888/0.110/72.74&28.00/0.797/0.194/106.6&31.18/0.908/0.091/57.00\\
				&IDR \cite{zhang2023ingredient}&28.68/0.968/0.018/18.55&35.99/0.968/0.025/23.65&33.89/0.932/0.063/39.13&\textbf{31.65}/\underline{0.889}/0.103/63.79&28.02/\underline{0.798}/0.187/102.6&31.64/0.911/0.079/49.54\\
				&PromptIR \cite{potlapalli2023promptir} &\underline{30.58}/\underline{0.974}/0.012/13.23&36.37/0.972/0.019/16.78&\underline{33.97}/\underline{0.933}/0.046/27.54&31.29/0.888/0.090/53.69&\underline{28.06}/\underline{0.798}/0.179/95.84&32.05/\underline{0.913}/0.069/41.42\\
				&DA-CLIP \cite{luocontrolling}&30.12/0.972/\underline{0.009}/\underline{8.952}&35.92/0.972/\underline{0.015}/13.73&33.86/0.925/\underline{0.045}/\underline{25.27}&31.06/0.865/\underline{0.082}/\underline{48.64}&27.55/0.778/0.168/89.28&31.70/0.901/\underline{0.063}/\underline{37.17}\\
				\cmidrule{2-8}
				& RCOT \cite{tang2024residualconditioned}  & 30.32/0.973/\underline{0.009}/10.52&\underline{37.25}/\underline{0.974}/\underline{0.015}/\underline{12.25}&33.86/0.932/0.048/30.12&31.20/0.886/0.096/57.25&28.03/0.797/\underline{0.162}/\underline{87.69}&\underline{32.13}/0.912/0.067/39.57\\
				&DA-RCOT & \textbf{31.26}/\textbf{0.977}/\textbf{0.007}/\textbf{4.058}&\textbf{38.36}/\textbf{0.983}/\textbf{0.008}/\textbf{6.154}& \textbf{33.98}/\textbf{0.934}/\textbf{0.038}/\textbf{21.05}&\underline{31.33}/\textbf{0.890}/\textbf{0.073}/\textbf{39.01} & \textbf{28.10}/\textbf{0.801}/\textbf{0.150}/\textbf{80.01} &\textbf{32.60}/\textbf{0.917}/\textbf{0.055}/\textbf{30.06}\\
				\bottomrule
		\end{tabular}}
		\vspace{-0.2cm}
	\end{table*}
	\begin{table*}[!h]
		\centering
		\caption{The \textit{All-in-One} comparison of our DA-RCOT with the state-of-the-art methods on \textbf{five} degradations. The metrics are reported in the form of PSNR($\uparrow$)/SSIM($\uparrow$)/LPIPS($\downarrow$)/FID($\downarrow$). ($*$) indicates the method is performed in an unpaired setting.}
		\label{air5}
		\vspace{-0.2cm}
		\setlength{\tabcolsep}{3pt}
		\renewcommand{\arraystretch}{1.2}
		\resizebox{\textwidth}{!}{
			\begin{tabular}{@{}clcccccc@{}}
				\toprule
				&\multirow{2}{*}{Method} 
				&\textit{Dehazing} & \textit{Deraining} & \textit{Denoising}&Deblurring&Low-light
				& \multirow{2}{*}{Average}  \\  \cmidrule(lr){3-3} \cmidrule(lr){4-4} \cmidrule(lr){5-5}  \cmidrule(lr){6-6}  \cmidrule(lr){7-7} 
				&& SOTS &Rain100L &  BSD68\textsubscript{$\sigma$=25} & GoPro& LOL-v1 &  \\
				\midrule
				\multirow{5}{*}{Unpaired}&CycleGAN$^*$\cite{zhu2017unpaired}    & 20.53/0.824/0.083/69.67 &  22.67/0.796/0.110/82.57  &   26.58/0.788/0.165/107.5&  18.25/0.650/0.256/102.2&   16.85/0.636/0.152/176.2 & 20.97/0.738/0.153/107.6  \\
				&WGAN-GP$^*$ \cite{gulrajani2017improved} & 20.92/0.856/0.076/58.52&30.45/0.912/0.087/54.28&27.67/0.792/0.168/108.9&19.19/0.665/0.221/94.23&17.27/0.640/0.140/168.1&23.10/0.773/0.138/96.82\\
				&OTUR$^*$ \cite{wang2022optimal}   & 21.41/0.871/0.072/57.82&30.86/0.919/0.086/53.84&28.01/0.804/0.165/108.4&19.78/0.671/0.220/93.00&17.49/0.649/0.136/167.1&23.51/0.783/0.136/96.03 \\
				\cmidrule{2-8}
				&RCOT$^*$ \cite{tang2024residualconditioned}   & \underline{24.55}/\underline{0.914}/\underline{0.052}/\underline{39.62}&\underline{32.83}/\underline{0.948}/\underline{0.062}/\underline{54.57}&\underline{28.65}/\underline{0.813}/\underline{0.147}/\underline{90.29}&\underline{21.36}/\underline{0.712}/\underline{0.206}/\underline{79.91}&\underline{18.73}/\underline{0.705}/\underline{0.127}/\underline{154.1}&\underline{25.22}/\underline{0.818}/\underline{0.118}/\underline{83.70}
				\\
				&DA-RCOT$^*$   & \textbf{24.87}/\textbf{0.923}/\textbf{0.041}/\textbf{30.28} & \textbf{33.19}/\textbf{0.953}/\textbf{0.054}/\textbf{45.72} & \textbf{28.73}/\textbf{0.815}/\textbf{0.140}/\textbf{83.20} & \textbf{21.85}/\textbf{0.727}/\textbf{0.188}/\textbf{63.26} & \textbf{19.21}/\textbf{0.722}/\textbf{0.118}/\textbf{123.9} & \textbf{25.57}/\textbf{0.828}/\textbf{0.108}/\textbf{69.27}
				\\
				\midrule
				\multirow{10}{*}{Paired}&MPRNet \cite{Zamir2021MPRNet}   & 24.28/0.931/0.061/43.55 &  33.12/0.927/0.064/57.84  &  30.18/0.846/0.112/83.47&  25.98/0.786/0.179/55.95&   18.98/0.776/0.115/103.5 & 26.51/0.853/0.106/68.86  \\
				&Restormer \cite{Zamir2021Restormer}   & 24.09/0.927/0.065/41.76 &  34.81/0.971/0.045/49.18  &  30.78/0.876/0.095/72.95&  27.22/0.829/0.174/56.10&   20.41/0.806/0.109/107.7 & 27.46/0.881/0.098/65.54 \\
				&IR-SDE \cite{luo2023image}   & 24.56/0.940/0.047/29.89 &  34.12/0.951/0.040/43.95  &  30.89/0.865/0.089/62.16
				&  26.34/0.800/0.162/48.77&   20.07/0.780/0.102/86.13 & 27.20/0.867/0.088/54.18\\
				\cmidrule{2-8}
				&DL \cite{fan2019general} & 20.54/0.826/0.096/76.25 & 21.96/0.762/0.122/87.15 & 23.09/0.745/0.155/154.2 & 19.86/0.672/0.218/85.48 & 19.83/0.712/0.120/125.8 &21.06/0.743/0.142/105.7\\
				&AirNet \cite{li2022all} &21.04/0.884/0.077/62.52 & 32.98/0.951/0.058/50.12 & 30.91/0.882/0.102/78.12 & 24.35/0.781/0.189/66.13 & 18.18/0.735/0.122/116.9&25.49/0.847/0.110/74.76\\
				&IDR \cite{zhang2023ingredient}&25.24/0.943/0.052/33.25 & 35.63/0.965/0.043/45.62 & \textbf{31.60}/\underline{0.887}/0.092/66.24 & 27.87/0.846/0.178/40.83 & 21.34/0.826/0.108/100.6&28.33/0.893/0.095/57.31\\
				&PromptIR \cite{potlapalli2023promptir} &\underline{30.41}/\underline{0.972}/0.017/20.12 & 36.17/0.970/0.024/22.53 & 31.20/0.885/0.097/66.91 & 27.93/0.851/0.155/29.52 & \underline{22.89}/0.829/0.098/70.32&29.72/0.901/0.078/41.88\\
				&DA-CLIP \cite{luocontrolling}&29.78/0.968/\underline{0.014}/\underline{15.26}&35.65/0.962/\underline{0.022}/22.24&30.93/0.885/\underline{0.089}/\underline{54.12}&27.31/0.838/\underline{0.143}/\underline{23.34}&21.66/0.828/\underline{0.095}/\underline{55.81}&29.07/0.896/\underline{0.073}/\underline{34.15}\\
				\cmidrule{2-8}
				&RCOT \cite{tang2024residualconditioned}& 30.26/0.971/0.016/16.74&\underline{36.88}/\underline{0.975}/0.024/\underline{19.67}& 31.05/0.882/0.099/62.12&\underline{28.12}/\underline{0.862}/0.155/21.56& 22.76/\underline{0.830}/0.097/61.24&\underline{29.81}/\underline{0.904}/0.078/36.26\\
				&DA-RCOT& \textbf{30.96}/\textbf{0.975}/\textbf{0.008}/\textbf{10.62}&\textbf{37.87}/\textbf{0.980}/\textbf{0.012}/\textbf{12.20}& \underline{31.23}/\textbf{0.888}/\textbf{0.082}/\textbf{37.65}&\textbf{28.68}/\textbf{0.872}/\textbf{0.135}/\textbf{12.39} & \textbf{23.25}/\textbf{0.836}/\textbf{0.084}/\textbf{47.23} &\textbf{30.40}/\textbf{0.911}/\textbf{0.064}/\textbf{24.02}\\
				\bottomrule
		\end{tabular}}
		\vspace{-0.4cm}
	\end{table*}
	\label{exp}
	We evaluate the proposed DA-RCOT on both real-world and synthetic datasets under two distinct configurations: 1) All-in-One and 2) Task-specific. In the All-in-One configuration, a single model is trained to handle multiple types of degradation, including three and five distinct degradation types on benchmark datasets. Conversely, the task-specific configuration requires the training of separate models, each dedicated to specific restoration tasks. We use PSNR/SSIM for measuring pixel-wise similarity, LPIPS \cite{zhang2018unreasonable}/FID \cite{heusel2017gans} for measuring perceptual deviation, and two well-known non-reference indexes NIQE \cite{mittal2012making} and PIQE  \cite{venkatanath2015blind} to assess real-world multiple-degradation images. The best and second-best are \textbf{highlighted} and \underline{underlined} respectively. The FID scores are calculated with 256$\times$256 center-cropped patches.
	
	\vspace{-0.2cm}
	\quad\\
	\noindent\textbf{Implementation details.} We train our models under all configurations using the RMSProp optimizer with a learning rate of $1\times10^{-4}$ for the transport network $T_\theta$ and $0.5\times10^{-4}$ for the potential network $\varphi_w$. The temperature hyperparameter $\tau$ is empirically set as 0.07. The inner iteration number $n_T$ is set to be 1.  In all experiments, the transport map $T_\theta$ is implemented with the backbone in Restormer \cite{Zamir2021Restormer} and the potential network is the same as \cite{ledig2017photo}. In the FROT objective, $c(y,x)$ is suggested as  $\|y-x\|$. During training, we crop patches of size 128x128 as input. All the experiments are conducted on Pytorch with an NVIDIA 4090 GPU.
	For the unpaired setting, although datasets that contain paired data are used for training, we randomly shuffle the target $x$ and degraded input $y$ to ensure the loss is isolated from paired information, which is a common strategy \cite{wang2022optimal, korotin2023kernel, korotin2023neural} for unpaired restoration.
	\quad\\
	\noindent\textbf{Datasets.} We evaluate the proposed method on benchmark \\datasets covering both synthetic and real-world data. For denoising, we merge  BSD400 \cite{arbelaez2010contour} and WED \cite{ma2016waterloo} datasets, adding Gaussian noise with levels $\sigma\in\{15,25,50\}$. Testing is conducted on the Kodak24 \cite{franzen1999kodak} and BSD68 \cite{martin2001database} datasets. For deraining, we use the Rain100L \cite{yang2017deep} and real-world SPANet \cite{Wang_2019_CVPR}. For dehazing, we employ SOTS \cite{li2018benchmarking} and real-world O-HAZE \cite{ancuti2018haze}. The deblurring and low-light enhancement tasks leverage GoPro \cite{nah2017deep} and LOL-v1 \cite{wei2018deep} datasets, respectively. For the All-in-One configuration, we merge these datasets into a mixed one with three or five degradation types for training a unified model.
	\begin{figure*}[!t]
		\setlength\tabcolsep{1pt}
		\renewcommand{\arraystretch}{0.7} 
		\centering
		\begin{tabular}{cccccc}
			\includegraphics[width=0.155\linewidth]{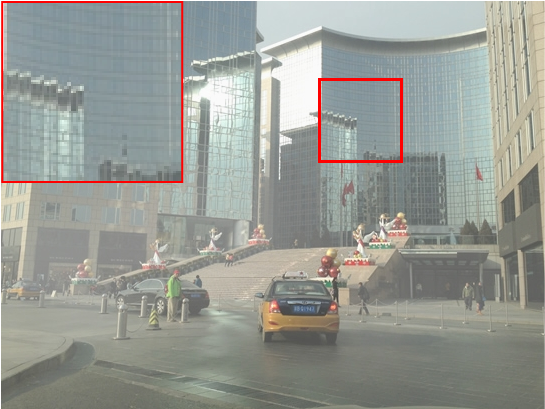}&
			\includegraphics[width=0.155\linewidth]{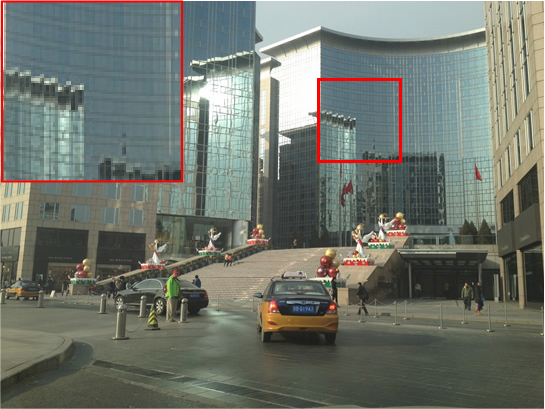}&
			\includegraphics[width=0.155\linewidth]{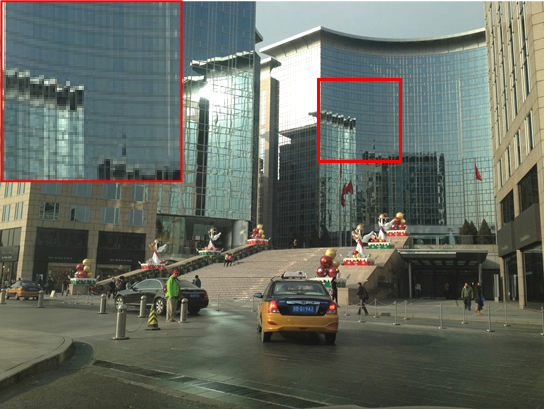}&
			\includegraphics[width=0.155\linewidth]{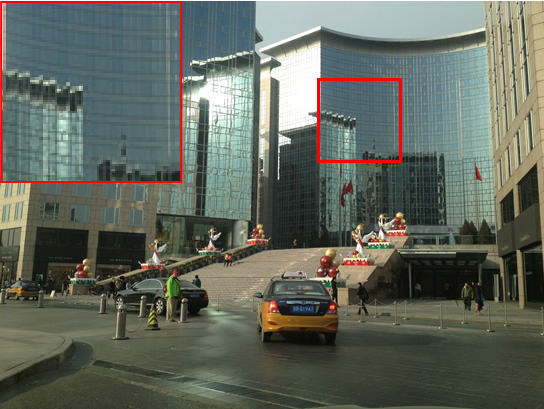}&
			\includegraphics[width=0.155\linewidth]{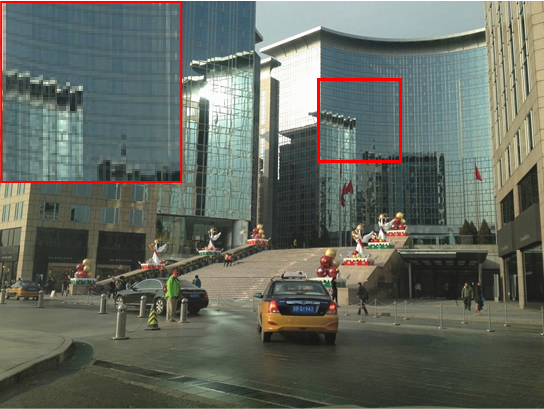}&
			\includegraphics[width=0.155\linewidth]{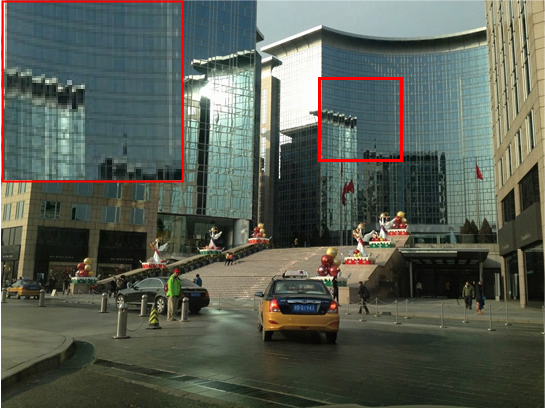}\\
			\includegraphics[width=0.155\linewidth]{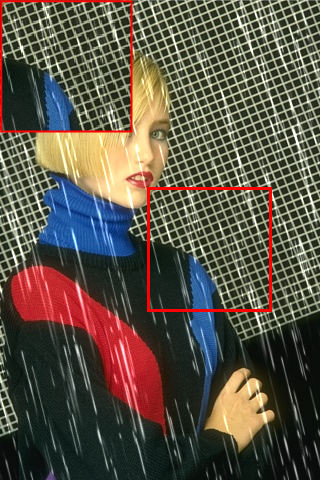}&
			\includegraphics[width=0.155\linewidth]{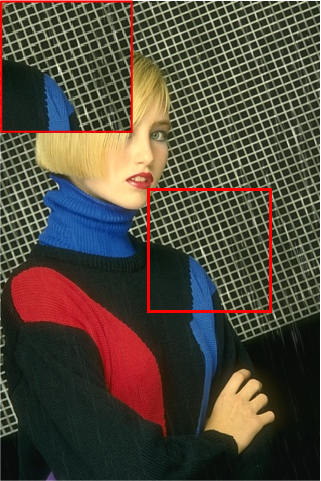}&
			\includegraphics[width=0.155\linewidth]{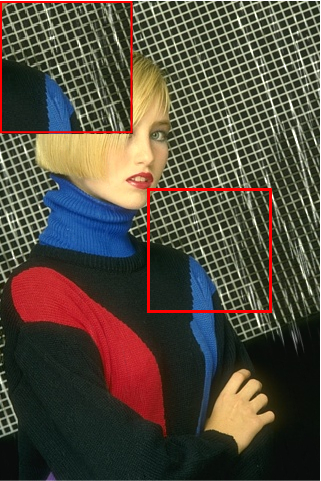}&
			\includegraphics[width=0.155\linewidth]{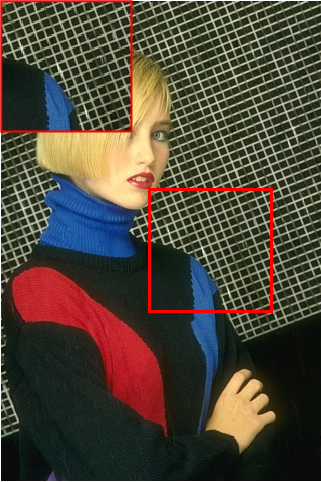}&
			\includegraphics[width=0.155\linewidth]{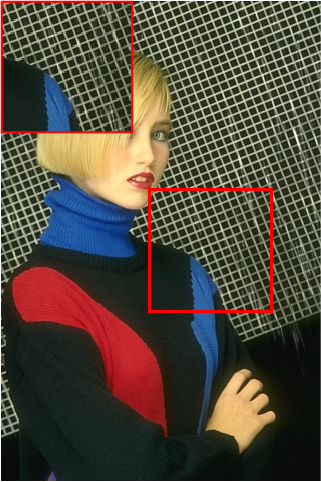}&
			\includegraphics[width=0.155\linewidth]{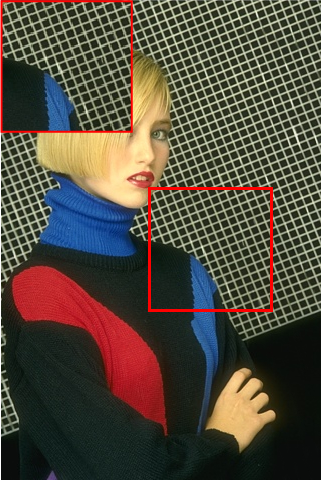}\\
			\includegraphics[width=0.155\linewidth]{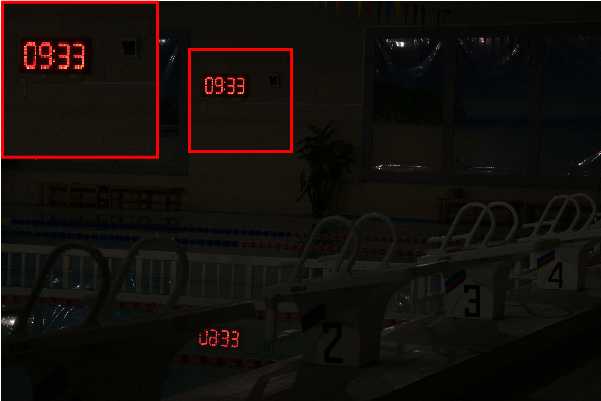}&
			\includegraphics[width=0.155\linewidth]{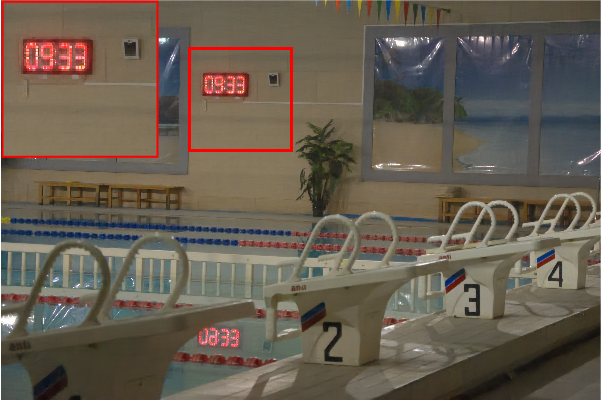}&
			\includegraphics[width=0.155\linewidth]{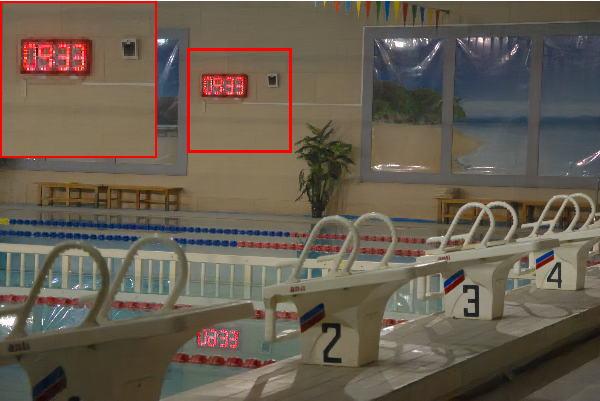}&
			\includegraphics[width=0.155\linewidth]{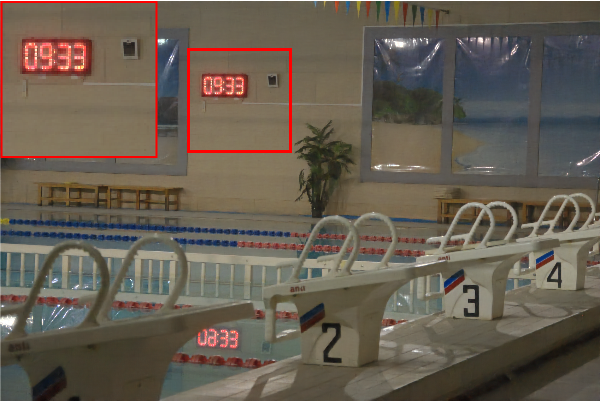}&
			\includegraphics[width=0.155\linewidth]{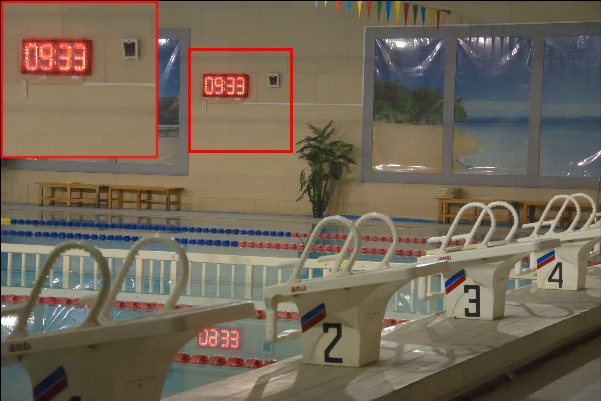}&
			\includegraphics[width=0.155\linewidth]{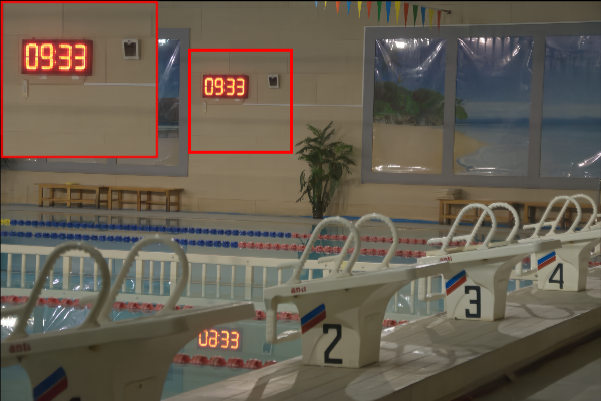}\\
			\includegraphics[width=0.155\linewidth]{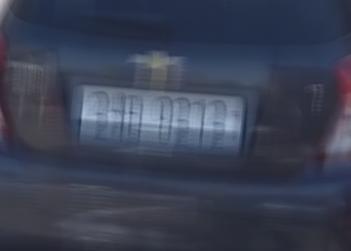}&
			\includegraphics[width=0.155\linewidth]{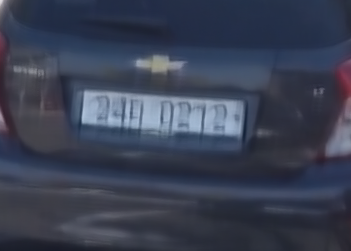}&
			\includegraphics[width=0.155\linewidth]{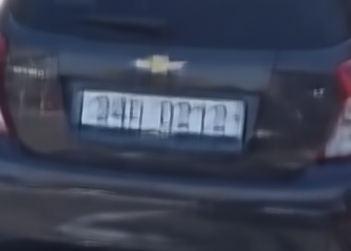}&
			\includegraphics[width=0.155\linewidth]{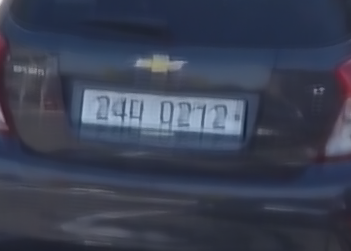}&
			\includegraphics[width=0.155\linewidth]{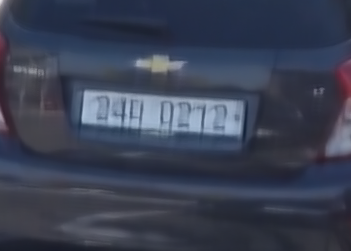}&
			\includegraphics[width=0.155\linewidth]{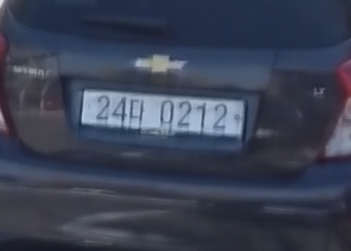}
			\\
			\includegraphics[width=0.155\linewidth]{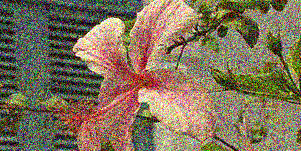}&
			\includegraphics[width=0.155\linewidth]{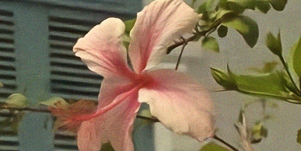}&
			\includegraphics[width=0.155\linewidth]{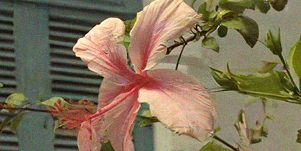}&
			\includegraphics[width=0.155\linewidth]{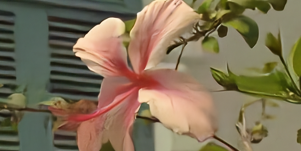}&
			\includegraphics[width=0.155\linewidth]{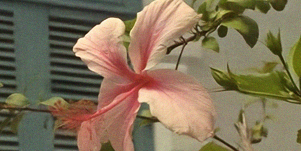}&
			\includegraphics[width=0.155\linewidth]{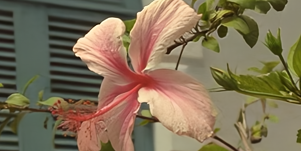}\\
			
			Degraded&Restormer&IR-SDE&PromptIR&DA-CLIP&DA-RCOT\\
		\end{tabular}
		\caption{Visual comparison of five-degradation All-in-One results. DA-RCOT restores sharp images with more faithful structural contents.}
		\label{vair5}
		\vspace{-0.3cm}
	\end{figure*} 
	\vspace{-0.2cm}
	\subsection{All-in-One Restoration Results}
	We compare our All-in-One DA-RCOT map with several state-of-the-art methods including three general restoration restorers, i.e., MPRNet \cite{Zamir2021MPRNet}, Restormer \cite{Zamir2021Restormer}, IR-SDE \cite{luo2023image}; and five specialized All-in-One models, i.e., DL \cite{fan2019general}, AirNet \cite{li2022all}, IDR \cite{zhang2023ingredient}, PromptIR \cite{potlapalli2023promptir}, DA-CLIP \cite{luocontrolling}. In particular, we also compete against the OT-based method and OTUR \cite{wang2022optimal} and the GAN-based method CycleGAN \cite{zhu2017unpaired} and WGAN-GP \cite{gulrajani2017improved} for restoration under an unpaired setting. 
	
	\textbf{Three degradations.} The first comparison is performed on three degradations, i.e., dehazing, deraining, and denoising on images of noise levels $\sigma\in\{15,25,50\}$. Table \ref{air3} reports the quantitative results, showing that our DA-RCOT performs best under all metrics among the competitors.  In the unpaired setting, our DA-RCOT outperforms the second-best method by an average of 2.11 dB in PSNR and 0.030 in LPIPS. In the paired setting, DA-RCOT  outpforms other Approaches in terms of both distortion measures (PSNR and SSIM) and perceptual quality measures (LPIPS and FID), offering an average improvement of 0.55 db in PSNR compared to the PromptIR \cite{potlapalli2023promptir}.  
	
	\textbf{Five degradations.} To evaluate the adaptability of DA-RCOT to more  tasks in All-in-One setting, we extend the comparison to a five-degradation scenario, adding the tasks of deblurring and low-light enhancement. As shown in Table \ref{air5}, DA-RCOT stands out with an average improvement of 0.68 dB over PromptIR \cite{potlapalli2023promptir} in PSNR and 10.13 reduction over DA-CLIP \cite{luocontrolling} in FID, which demonstrates the adaptability of the DA-RCOT map for more diverse degradations.
	
	Figure \ref{vair5} displays the visual results of the evaluated methods under five degradation scenarios. PromptIR produces results results with over-smoothed structures or artifacts. DA-CLIP \cite{luocontrolling} can produce seemingly realistic results but with remaining distortion like blur and noise. In comparison, our DA-RCOT can effectively remove the degradation while preserving the image structures.
	\begin{table*}[!t]
		\centering
		\caption{The \textit{Task-specific} comparison on single degradations. The metrics are reported in the form of PSNR($\uparrow$)/SSIM($\uparrow$)/LPIPS($\downarrow$)/FID($\downarrow$).	}
		\vspace{-0.3cm}
		\label{single}
		\begin{subtable}[h]{0.46\textwidth}
			\subcaption{\textit{Deraining}}
			\vspace{-0.2cm}
			\setlength{\tabcolsep}{5pt}
			\resizebox{\textwidth}{!}{
				\begin{tabular}{lcc}
					\toprule
					Method &Synthetic Rain100L &Real-world SPANet\\
					\midrule
					MPRNet \cite{Zamir2021MPRNet}& 34.95/0.964/0.039/21.61 & 39.52/0.967/0.021/28.13\\
					Restormer \cite{Zamir2021Restormer}& 36.74/0.978/0.026/13.29 & 41.39/0.981/0.013/19.67 \\
					SFNet \cite{cui2023selective}& 36.56/0.974/0.023/13.12 & 41.02/0.980/0.015/21.52\\
					IR-SDE \cite{luo2023image}&36.94/0.978/0.014/9.521 &42.43/0.986/0.013/17.25\\
					\midrule
					AirNet \cite{li2022all} & 34.90/0.977/0.035/20.96&39.67/0.969/0.019/25.62\\
					PromptIR \cite{potlapalli2023promptir} &37.23/0.980/0.017/10.21&41.17/0.979/0.019/21.52  \\
					DA-CLIP \cite{luocontrolling}&37.02/0.978/\underline{0.012}/8.963 &42.84/0.988/\underline{0.009}/14.61\\
					\midrule
					RCOT \cite{tang2024residualconditioned}&\underline{37.73}/\underline{0.981}/0.013/\underline{7.551} & \underline{43.98}/\underline{0.994}/0.011/\underline{8.457}\\
					DA-RCOT &\textbf{38.73}/\textbf{0.985}/\textbf{0.007}/\textbf{5.013} &\textbf{45.33}/\textbf{0.997}/\textbf{0.006}/\textbf{6.289}\\
					\bottomrule
				\end{tabular}
			}
		\end{subtable}%
		\begin{subtable}[h]{0.48\textwidth}
			\subcaption{\textit{Dehazing}}
			\vspace{-0.2cm}
			\setlength{\tabcolsep}{5pt}
			\resizebox{\textwidth}{!}{
				\begin{tabular}{lcc}
					\toprule
					Method &Synthetic SOTS &Real-world O-HAZE\\
					\midrule
					MPRNet \cite{Zamir2021MPRNet}&28.31/0.954/0.029/17.79 & 21.55/0.778/0.256/223.8\\
					Restormer \cite{Zamir2021Restormer} &30.87/0.969/0.026/13.29&25.20/0.804/0.221/198.0\\
					IR-SDE \cite{luo2023image}&30.55/0.968/0.018/12.76 & 22.13/0.776/0.160/179.2\\ 
					Dehazeformer \cite{song2023vision}& 31.45/0.978/0.021/15.54&25.56/0.812/0.209/199.3\\
					\midrule
					AirNet \cite{li2022all} & 28.52/0.956/0.027/17.22&21.42/0.774/0.231/219.3\\
					PromptIR \cite{potlapalli2023promptir} &31.31/0.973/0.021/16.28&25.27/0.813/0.216/217.6 \\
					DA-CLIP \cite{luocontrolling}&30.88/0.974/\underline{0.009}/\underline{8.657}&23.86/0.782/0.158/\underline{163.8}\\
					\midrule
					RCOT \cite{tang2024residualconditioned}& \underline{31.52}/\underline{0.976}/0.014/10.05 & \underline{27.10}/\underline{0.838}/\underline{0.155}/167.7 \\ 
					DA-RCOT &\textbf{31.89}/\textbf{0.978}/\textbf{0.007}/\textbf{4.058}&\textbf{27.78}/\textbf{0.845}/\textbf{0.140}/\textbf{148.5}\\
					\bottomrule
				\end{tabular}
			}
		\end{subtable}
		\\
		\vspace{0.2cm}
		\begin{subtable}[h]{0.94\textwidth}
			\subcaption{\textit{Denoising}}
			\setlength{\tabcolsep}{3pt}
			\vspace{-0.2cm}
			\resizebox{\textwidth}{!}{
				\begin{tabular}{lcccccc}
					\toprule
					\multirow{2}{*}{Method} &\multicolumn{3}{c}{Kodak24} & \multicolumn{3}{c}{BSD68}\\ \cmidrule(lr){2-4} \cmidrule(lr){5-7}
					&$\sigma=15$&   $\sigma = 25$ & $\sigma = 50$ &$\sigma=15$&  $\sigma = 25$ & $\sigma = 50$ \\
					\midrule
					MPRNet \cite{Zamir2021MPRNet}&34.26/0.901/0.072/41.35 &31.96/0.868/0.112/43.98&28.36/0.785/0.185/73.26&33.74/0.931/0.076/39.34 &30.89/0.880/0.103/59.23&27.73/0.782/0.195/103.6\\
					Restormer\cite{Zamir2021Restormer}&34.73/0.918/0.060/24.76 &32.13/0.880/0.097/33.22&29.25/0.799/0.156/64.26&34.13/0.936/0.049/25.19 &31.60/0.895/0.085/49.28&28.35/0.801/0.177/88.26\\
					IR-SDE \cite{luo2023image}&33.92/0.898/0.068/30.11 &31.40/0.842/0.089/45.56&28.03/0.721/0.145/73.66&33.14/0.907/0.050/28.22 &30.46/0.856/0.086/46.30&26.98/0.737/0.172/86.40\\
					RCD\cite{Zhang_2023_CVPR}&34.56/0.916/0.070/39.21&32.18/0.880/0.089/39.85&29.13/0.795/0.157/63.26&33.98/0.934/0.065/36.77&31.28/0.886/0.089/50.96&28.01/0.796/0.186/91.78\\
					\midrule
					AirNet \cite{li2022all} &34.51/0.915/0.076/35.24&31.93/0.867/0.099/32.78&28.70/0.793/0.176/77.12&34.02/0.934/0.068/38.14 &31.48/0.893/0.094/55.12&28.11/0.806/0.184/93.94\\
					PromptIR \cite{potlapalli2023promptir}&34.85/0.926/\underline{0.054}/17.52& 32.25/0.883/0.091/30.91 & 29.49/\underline{0.809}/0.154/60.42& \underline{34.26}/\underline{0.937}/0.046/24.77& \textbf{31.71}/0.897/0.085/45.45&\underline{28.49}/\underline{0.813}/0.172/85.46\\
					DA-CLIP \cite{luocontrolling}&34.43/0.905/0.065/\underline{16.28}&31.78/0.865/\underline{0.079}/28.24&
					28.73/0.775/0.144/65.96&34.05/0.933/\textbf{0.033}/\underline{23.75}&31.20/0.883/\underline{0.072}/\underline{43.51}&27.85/0.788/0.162/84.22\\
					\midrule
					RCOT \cite{tang2024residualconditioned}&\underline{34.92}/\underline{0.928}/0.059/18.95&  \underline{32.68}/\textbf{0.885}/0.083/\underline{24.33} & \underline{29.53}/0.808/\underline{0.132}/\underline{48.54} &34.19/0.935/0.046/26.51& 31.65/0.896/0.075/48.24 & 28.31/0.808/\underline{0.160}/\underline{80.86} \\ 
					DA-RCOT&\textbf{35.02}/\textbf{0.929}/\textbf{0.048}/\textbf{13.15}&\textbf{32.75}/\textbf{0.885}/\textbf{0.072}/\textbf{18.15}&\textbf{29.65}/\textbf{0.810}/\textbf{0.121}/\textbf{40.75}&\textbf{34.29}/\textbf{0.939}/\underline{0.036}/\textbf{16.15}&\underline{31.70}/\textbf{0.898}/\textbf{0.060}/\textbf{38.87}&\textbf{28.59}/\textbf{0.816}/\textbf{0.145}/\textbf{72.27}\\
					\bottomrule
				\end{tabular}
			}
		\end{subtable}%
		\vspace{-0.1cm}
	\end{table*}
	\vspace{-0.2cm}
	\subsection{Task-specific Restoration Results}
	\begin{figure*}[!t]
		\setlength\tabcolsep{1pt}
		\renewcommand{\arraystretch}{0.618} 
		\centering
		\begin{tabular}{cccccc}
			\includegraphics[width=0.155\linewidth]{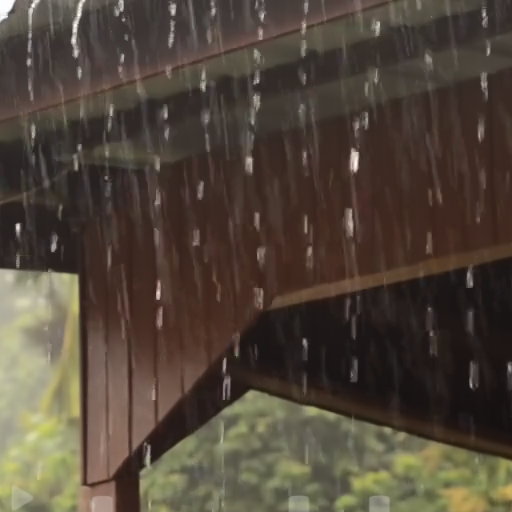}&
			\includegraphics[width=0.155\linewidth]{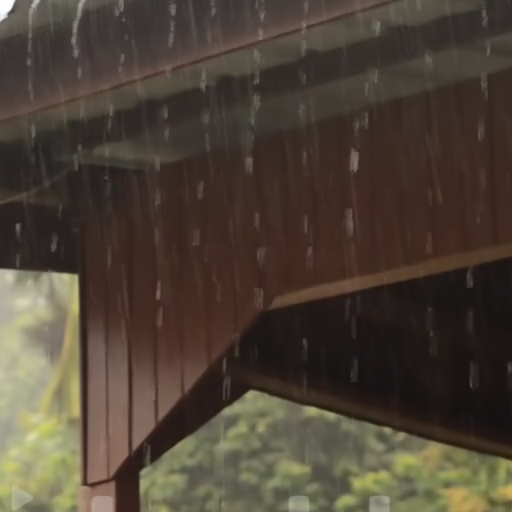}&
			\includegraphics[width=0.155\linewidth]{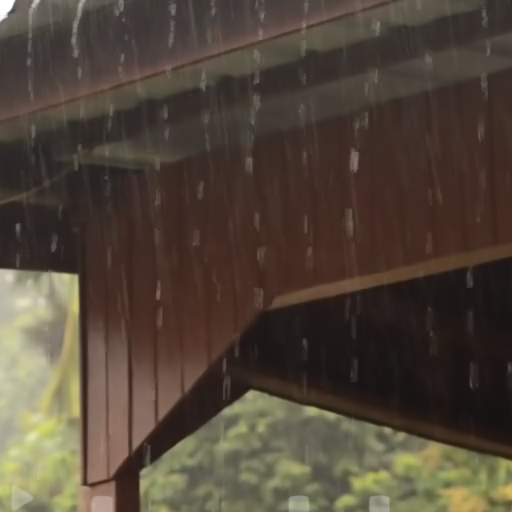}&
			\includegraphics[width=0.155\linewidth]{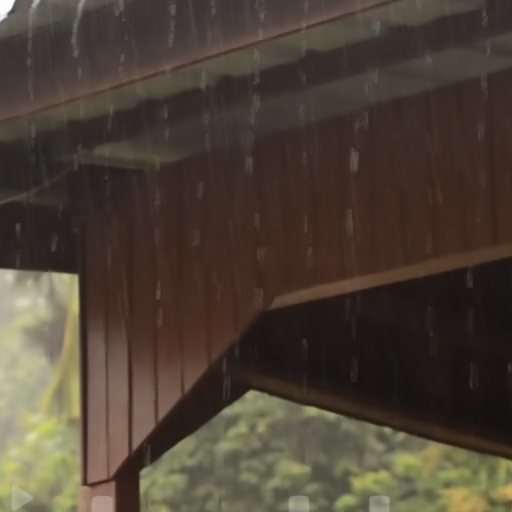}&
			\includegraphics[width=0.155\linewidth]{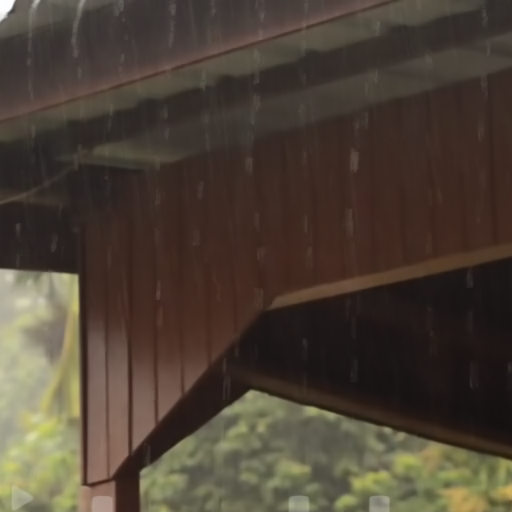}&
			\includegraphics[width=0.155\linewidth]{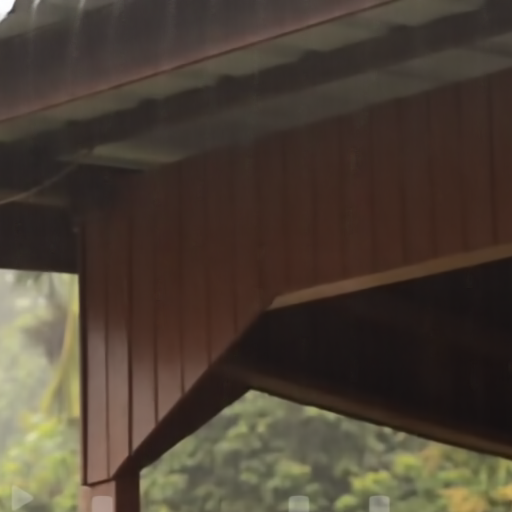}\\
			\includegraphics[width=0.155\linewidth]{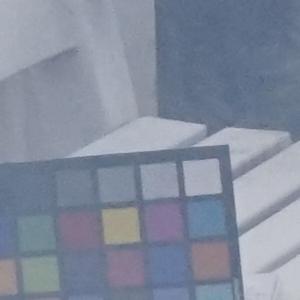}&
			\includegraphics[width=0.155\linewidth]{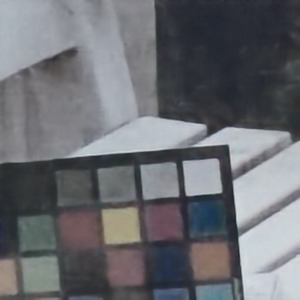}&
			\includegraphics[width=0.155\linewidth]{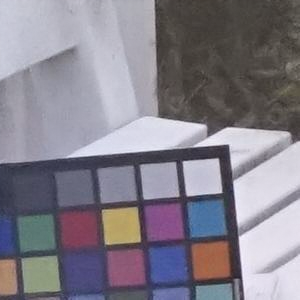}&
			\includegraphics[width=0.155\linewidth]{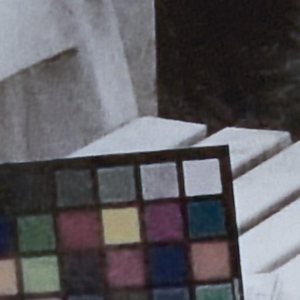}&
			\includegraphics[width=0.155\linewidth]{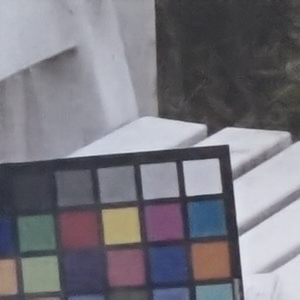}&
			\includegraphics[width=0.155\linewidth]{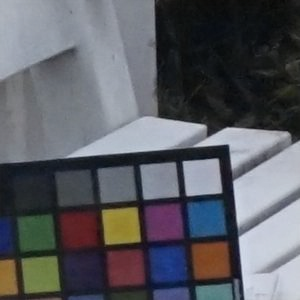}\\
			\includegraphics[width=0.155\linewidth]{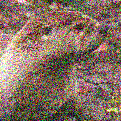}&
			\includegraphics[width=0.155\linewidth]{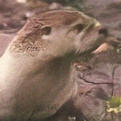}&
			\includegraphics[width=0.155\linewidth]{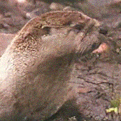}&
			\includegraphics[width=0.155\linewidth]{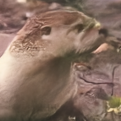}&
			\includegraphics[width=0.155\linewidth]{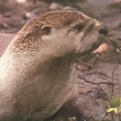}&
			\includegraphics[width=0.155\linewidth]{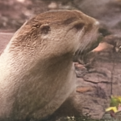}\\
			Degraded&Restormer&IR-SDE&PromptIR&DA-CLIP&DA-RCOT\\
		\end{tabular}
		\caption{Visual comparison of task-specific results of a real hazy from O-HAZE, a real rainy image from SPANet, and a noisy image from BSD68 with $\sigma=50$. DA-RCOT restores sharp images with more faithful structural contents (colors and textures). Please zoom in for better visualization.}
		\label{vspec}
		\vspace{-0.4cm}
	\end{figure*} 
	We also validate the effectiveness of DA-RCOT under task-specific configuration by training individual models over the single-degradation datasets. We evaluate the approaches for comparison on Rain100L \cite{yang2017deep} and real-world SPANet \cite{Wang_2019_CVPR} for deraining, SOTS \cite{li2018benchmarking} and real-world O-HAZE \cite{ancuti2018haze} for dehazing, BSD68 \cite{martin2001database} for denoising.  For each task, we add a task-specific state-of-the-art method for each comparison, i.e., SFNet \cite{cui2023selective} for deraining, Dehazeformer \cite{song2023vision} for dehazing, and RCD \cite{Zhang_2023_CVPR} for denoising. 	
	
	Table \ref{single} reports the quantitative results of the restoration methods under task-specific configuration. Across all tasks, DA-RCOT consistently outperforms other approaches in all metrics. Specifically, in terms of PSNR, DA-RCOT surpasses PromptIR \cite{potlapalli2023promptir} by 1.94 dB on Rain100L and outperforms DA-CLIP \cite{luocontrolling} by 2.49 dB on SPANet. When compared to the highly competitive task-specific Dehazeformer \cite{song2023vision} for dehazing, DA-RCOT achieves an improvement of 0.44 dB on SOTS and 2.22 dB on O-HAZE. The visual resultsParticularly, as compared to the preliminary work RCOT \cite{tang2024residualconditioned}, DA-RCOT still offers a non-trivial improvement. Figure \ref{vspec} displays three task-specific examples, where DA-RCOT produces cleaner results with more faithful structures (textures and colors). These results further underscore its expressive power for addressing single degradations in comparison to other all-in-one models and general restorers.
	\vspace{-0.2cm}
	\subsection{Robustness to the Number of Degradation Types}
	\begin{figure*}[!t]
		\centering
		\includegraphics[scale=0.55]{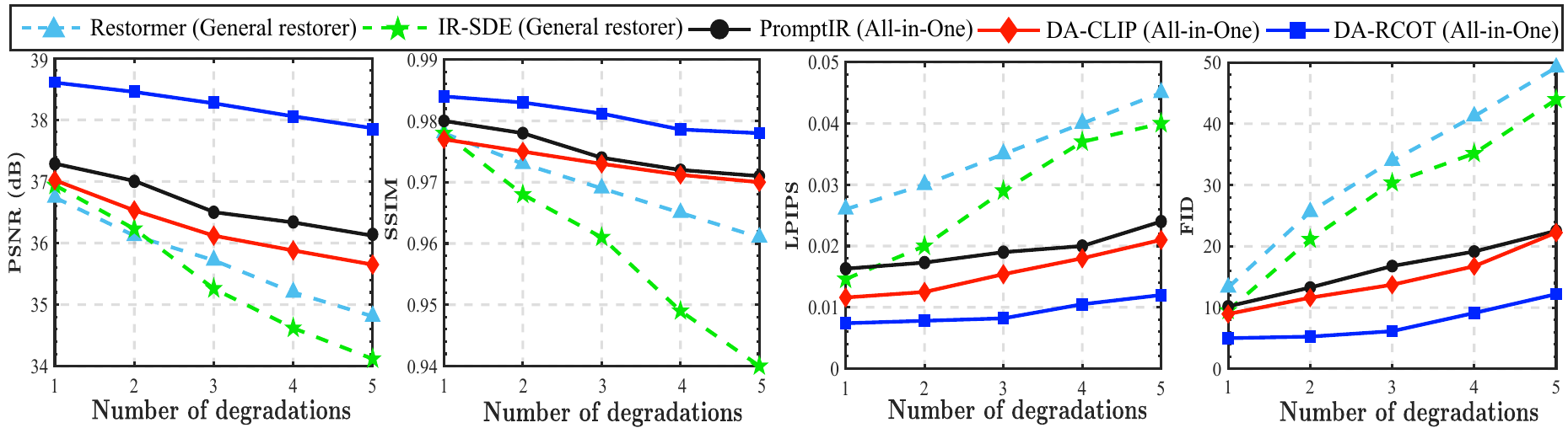}
		\vspace{-0.3cm}
		\caption{Numerical comparison of robustness to the number of degradation types. As the degradation number increases, DA-RCOT delivers pleasant robustness while achieving the best quantitative performance on the deraining task in comparison with other methods.}
		\label{Rob}
		\vspace{-0.4cm}
	\end{figure*}
	The robustness to degradation number is an important property of All-in-One models. Specifically, how well can these models adapt as the number of degradations increases? To explore this capability, we establish a task queue: 1) deraining, 2) denoising with $\sigma=25$, 3) dehazing, 4) deblurring, and 5) low-light enhancement. The tasks are added sequentially for all-in-one training and we record the draining performance on Rain100L  as the degradation number increases. The dataset settings follow those of the five-degradation All-in-One evaluation. 
	
	\noindent\textbf{Compared Methods.} Four methods are selected for comparisons, including general restorers Restormer \cite{Zamir2021Restormer}, IR-SDE \cite{luo2023image} and All-in-One PromptIR \cite{potlapalli2023promptir} and DA-CLIP \cite{luocontrolling}.  
	
	Figure \ref{Rob} displays the quantitative results of deraining against the number of degradation types. We can observe from the figure that as compared to the general restorers \cite{Zamir2021MPRNet}, \cite{Zamir2021Restormer}, the All-in-One models are generally more robust as the number of degradations increases. Notably, DA-RCOT achieves the highest quantitative performance and exhibits the least decline as the number of degradations goes up, which demonstrates its better robustness as compared to other methods. The underlying reason should be that the residual embeddings of DA-RCOT encode more intrinsic degradation semantics, dynamically adapting the OT map to handle heterogeneous degradations.
	\subsection{Generalization to Unseen Degradation Levels}	
	\begin{table}[!h]
		\vspace{-0.2cm}
		\centering
		\caption{The OOD deraining results on Rain100L (light rain images), where the models are trained on Rain100H (heavy rain images).}
		\label{unseen1}
		\vspace{-0.2cm}
		\resizebox{1.0\linewidth}{!}{
			\begin{tabular}{lcccc}
				\toprule
				Method & PSNR ($\uparrow$) & SSIM ($\uparrow$) & LPIPS ($\downarrow$)& FID ($\downarrow$)\\
				\midrule
				Restormer \cite{Zamir2021Restormer}   & 29.85&0.921&0.121&53.54 \\
				IR-SDE \cite{luo2023image} &29.53 & 0.912&0.092 &33.65\\
				\midrule
				AirNet \cite{li2022all} & 30.02	&0.933&0.110&46.19\\
				PromptIR \cite{potlapalli2023promptir} & 31.65&0.929&0.076&43.41 \\
				DA-CLIP \cite{luocontrolling}&\underline{33.61}&\underline{0.950}&0.056&34.11\\
				\midrule
				RCOT \cite{tang2024residualconditioned}  & 32.85&0.945&\underline{0.048}&\underline{32.63}\\
				DA-RCOT &\textbf{34.23}&\textbf{0.962}&\textbf{0.032}&\textbf{16.77}\\
				\bottomrule
		\end{tabular}}
		\vspace{0.2cm}
		\centering
		\caption{The OOD denoising results on BSD68 with unseen noise levels $\sigma=60$ and $\sigma=75$, where the models are trained on noisy images from BSD400 \cite{arbelaez2010contour} and WED \cite{ma2016waterloo} with noise levels $\sigma\in\{15,25,50\}$.}
		\label{unseen2}
		\vspace{-0.3cm}
		\resizebox{1.0\linewidth}{!}{
			\begin{tabular}{lcc}
				\toprule
				Method & $\sigma=60$ & $\sigma=75$\\
				\midrule
				Restormer \cite{Zamir2021Restormer}   &19.23/0.481/0.242/117.7&15.56/0.398/0.423/165.6\\
				IR-SDE \cite{luo2023image} &17.84/0.413/0.236/122.0&15.10/0.337/0.375/148.3\\
				\midrule
				AirNet \cite{li2022all} & 19.25/0.483/0.249/118.6	&15.87/0.406/0.392/164.6\\
				
				PromptIR \cite{potlapalli2023promptir} & 23.78/0.616/0.223/101.2&19.23/0.502/0.389/152.3\\
				DA-CLIP \cite{luocontrolling}&20.18/0.495/0.198/93.15&16.78/0.384/0.342/133.2\\
				\midrule
				RCOT \cite{tang2024residualconditioned}& \underline{25.12}/\underline{0.656}/\underline{0.184}/\underline{88.13}&\underline{19.85}/\underline{0.515}/\underline{0.326}/\underline{121.5}\\
				DA-RCOT &\textbf{25.89}/\textbf{0.685}/\textbf{0.167}/\textbf{66.95}&\textbf{21.64}/\textbf{0.571}/\textbf{0.276}/\textbf{108.6}\\
				\bottomrule
		\end{tabular}}
	\end{table}
	As observed in Figure \ref{emd_vis}, the residual embeddings of different noise levels are clustered together, exhibiting level-specific positional relationships. To validate the advantages of these characteristics, we evaluate DA-RCOT on out-of-distribution (OOD) data with unseen degradation levels, where the model was trained on data with other levels of the same degradation. Specifically, we conduct two experiments. In the first evaluation, the DA-RCOT map is trained on Rain100H \cite{yang2017deep} (heavily rainy images) and tested on Rain100L \cite{yang2017deep} (lightly rainy images). In the second evaluation, we test the performance of the DA-RCOT map on severe noise levels $\sigma=60$ and $\sigma=75$, where the model is trained on the mixed noisy images from BSD400 \cite{arbelaez2010contour} and WED \cite{ma2016waterloo} datasets with noise levels $\sigma\in\{15,25,50\}$.
	
	Tables \ref{unseen1} and \ref{unseen2} present the deraining and denoising quantitative results on OOD degradation levels. We can observe that All-in-One models generally perform better than regular general restorers. DA-RCOT exhibits superior generalization ability to unseen degradation levels as compared to other methods, highlighting the advantages of the residual embedding in capturing the intrinsic degradation semantics beyond levels, which coincides with the unique cluster patterns observed in Figure \ref{emd_vis}. Different from RCOT, DA-RCOT disentangles multi-scale embeddings and conditions them at different scales within the transport map network. This approach imparts richer correlations between levels and more intrinsic degradation semantics, thereby enhancing the OT map's generalization ability across degradation levels.
	\vspace{-0.2cm}
	\subsection{Real-world Images with Multiple Degradations}
	\begin{figure}[!h]
		\setlength\tabcolsep{0.3pt}
		\renewcommand{\arraystretch}{0.3} 
		\centering
		\begin{tabular}{cccc}
			\includegraphics[width=0.24\linewidth]{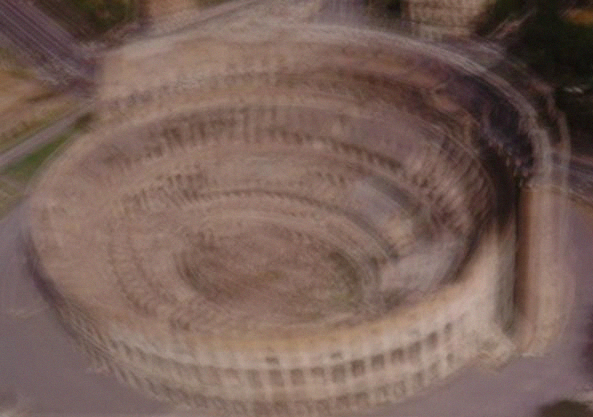}&
			\includegraphics[width=0.24\linewidth]{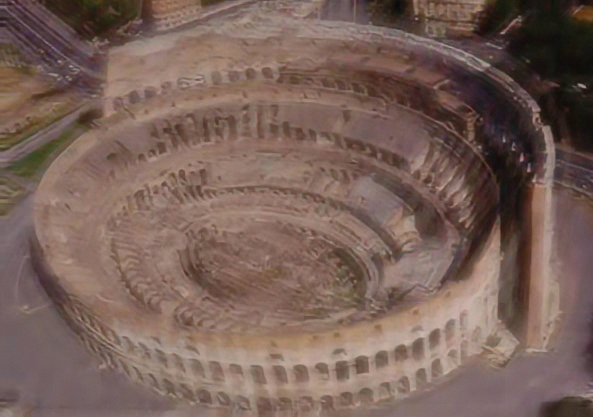}&
			\includegraphics[width=0.24\linewidth]{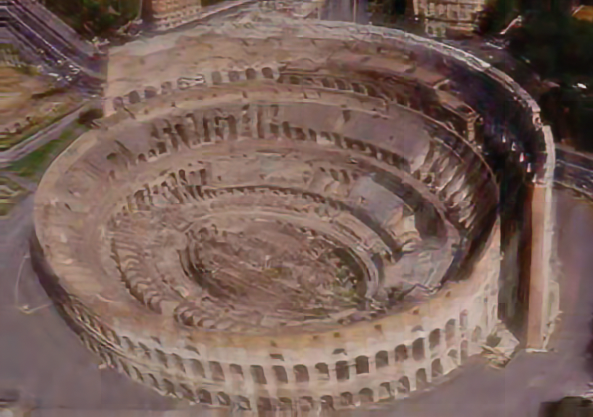}&
			\includegraphics[width=0.24\linewidth]{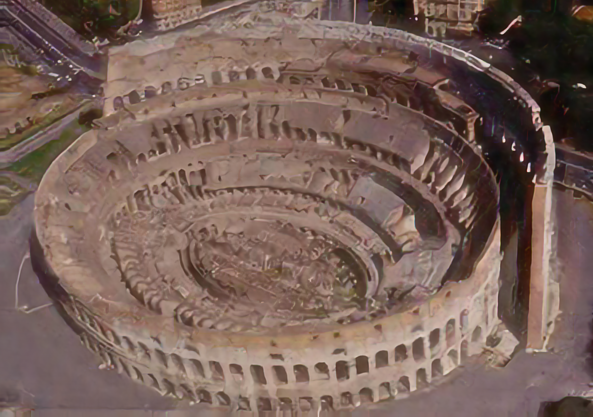}\\
			\includegraphics[width=0.24\linewidth]{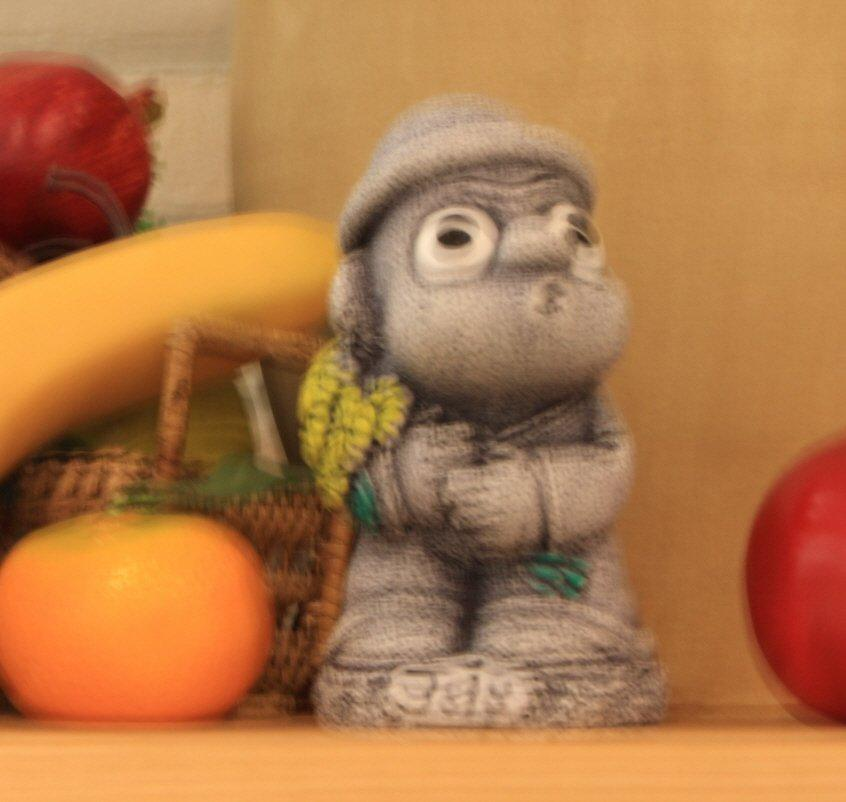}&
			\includegraphics[width=0.24\linewidth]{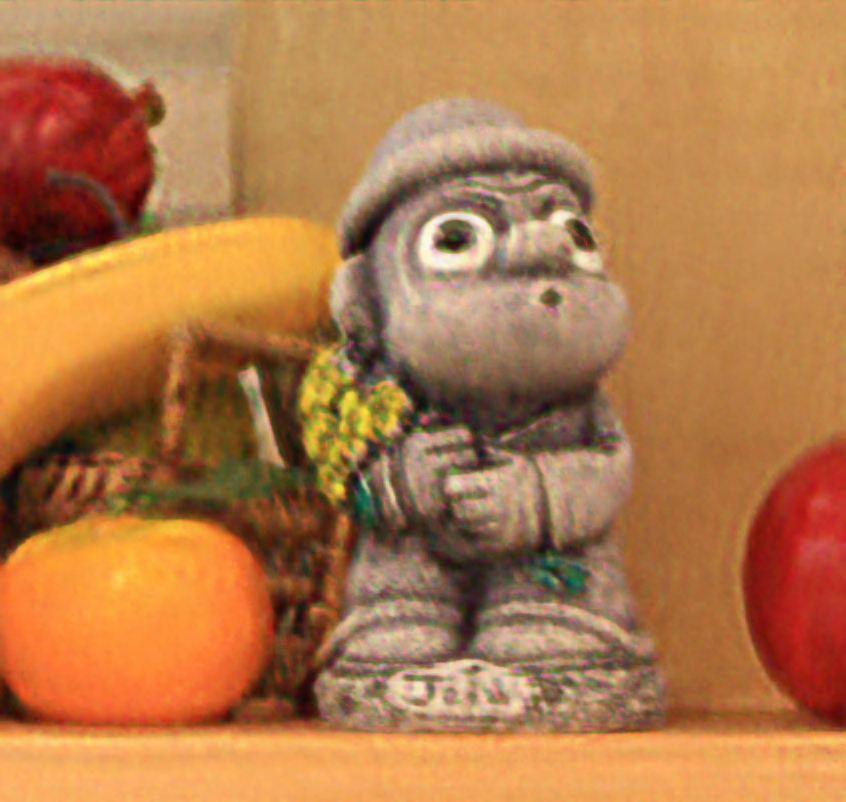}&
			\includegraphics[width=0.24\linewidth]{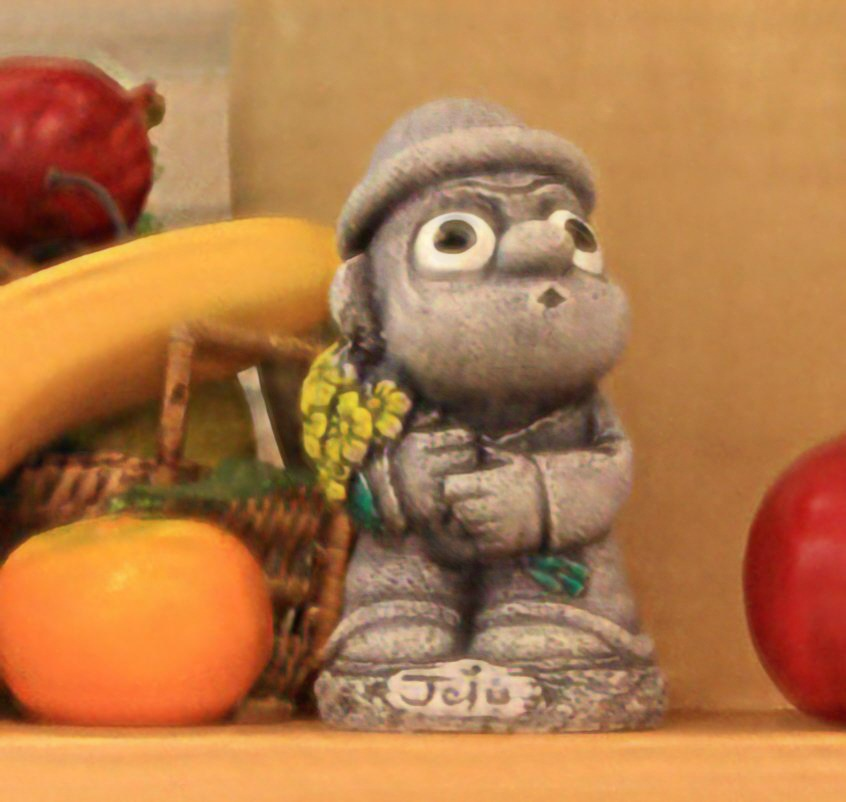}&
			\includegraphics[width=0.24\linewidth]{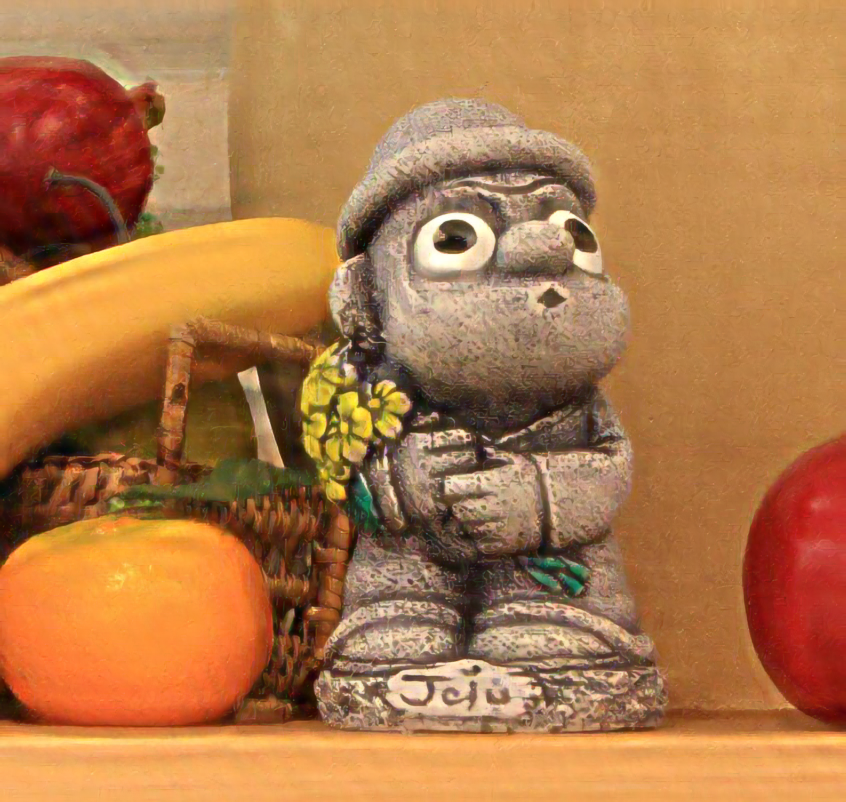}\\
			Blur + Noise&DA-CLIP&PromptIR&DA-RCOT\\
			\includegraphics[width=0.24\linewidth]{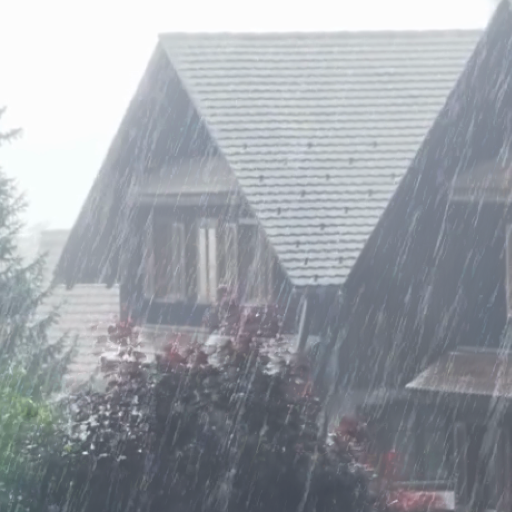}&
			\includegraphics[width=0.24\linewidth]{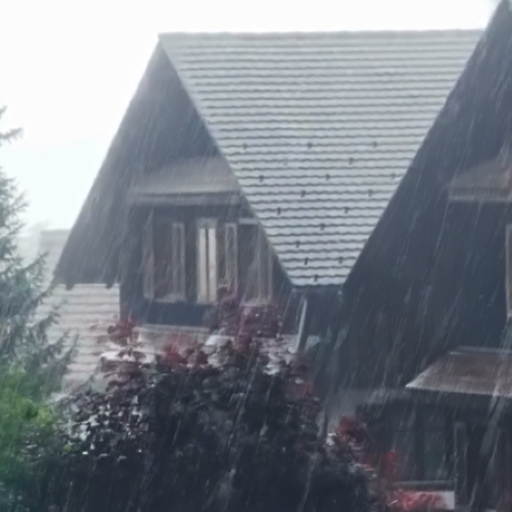}&
			\includegraphics[width=0.24\linewidth]{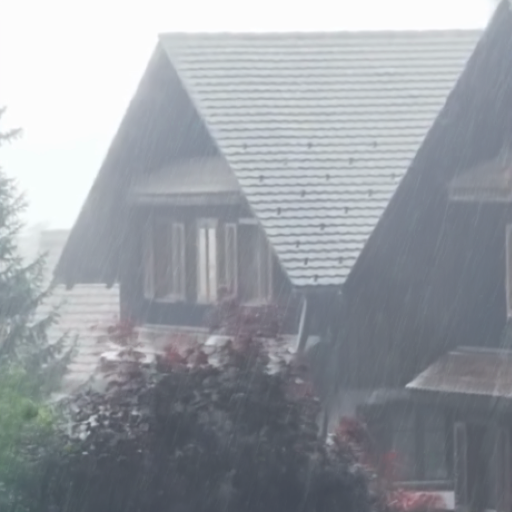}&
			\includegraphics[width=0.24\linewidth]{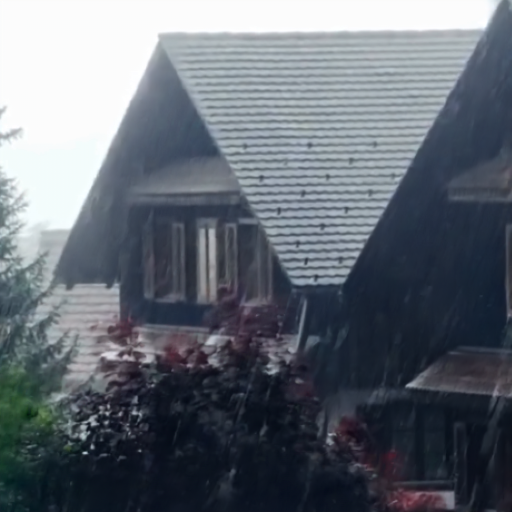}\\
			\includegraphics[width=0.24\linewidth]{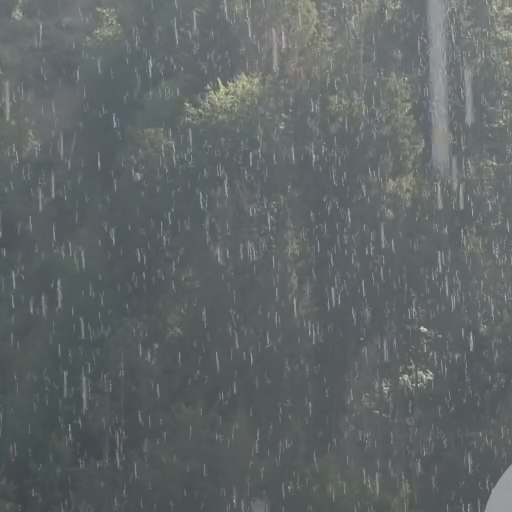}&
			\includegraphics[width=0.24\linewidth]{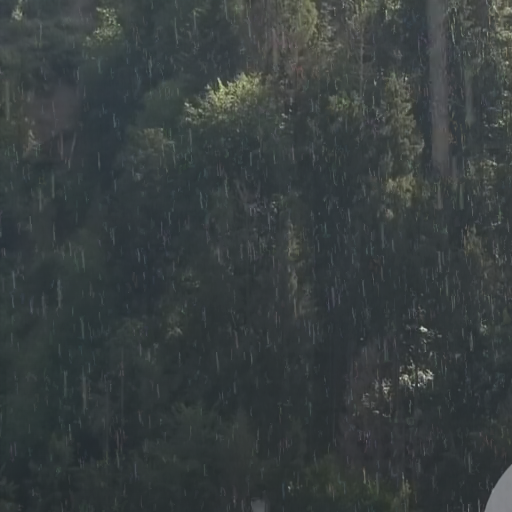}&
			\includegraphics[width=0.24\linewidth]{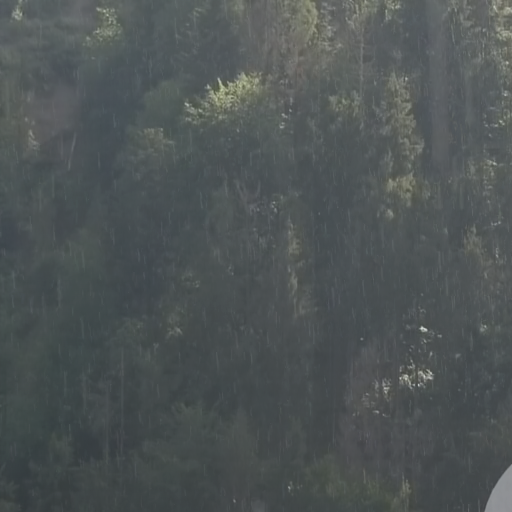}&
			\includegraphics[width=0.24\linewidth]{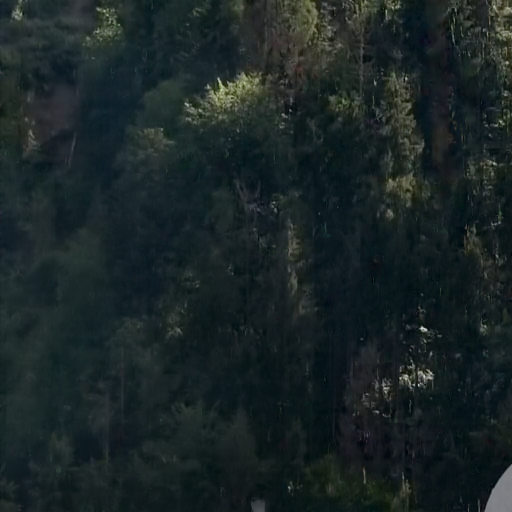}\\
			Haze + Rain&DA-CLIP&PromptIR&DA-RCOT
		\end{tabular}
		\caption{Visual comparisons of All-in-One models addressing multiple degradations. The first two rows display the results for blurry and noisy images, while the last two rows display those for hazy and rainy images.}
		\label{multi-deg}
	\end{figure}
	\begin{table*}[!t]
		\centering
		\caption{The qualitative results of the model with different conditions for the second pass restoration. Metrics are reported in the form of PSNR($\uparrow$)/LPIPS($\downarrow$). }
		\label{T-REC}
		\vspace{-0.2cm}
		\setlength{\tabcolsep}{12pt}
		\renewcommand{\arraystretch}{1}
		\resizebox{\textwidth}{!}{
			\begin{tabular}{l|ccccc|c}
				\toprule
				Method& SOTS &Rain100L &  BSD68\textsubscript{$\sigma$=25} & GoPro& LOL &Average  \\
				\midrule
				w/o any condition  									
				&27.56/0.054 &35.37/0.038&31.10/0.105&27.32/0.177&20.60/0.135&28.39/0.102\\
				conditioned on $\hat x_0$ &27.87/0.052&35.82/0.039&30.43/0.115&27.41/0.172&20.73/0.122&28.45/0.100\\
				conditioned on $\mathbf R_0$ & 30.26/0.016 &36.88/0.024&31.05/0.099&28.12/0.155&22.76/0.097&29.81/0.078\\
				conditioned on $\{\mathbf R_i\}_{i=1,2,3}$ & \textbf{30.96}/\textbf{0.008}&\textbf{37.87}/\textbf{0.012}&\textbf{31.22}/\textbf{0.082}&\textbf{28.68}/\textbf{0.135}&\textbf{23.25}/\textbf{0.084}&\textbf{30.40}/\textbf{0.064}\\
				\bottomrule
		\end{tabular}}
		\vspace{-0.3cm}
	\end{table*}
	The images captured from the real world are sometimes corrupted by multiple degradations due to adverse weather conditions and the property of imaging devices. In this sense, a practical challenge for the All-in-One models is to handle these multiple degradations. Here, we compare DA-RCOT with PromptIR \cite{potlapalli2023promptir} and DA-CLIP \cite{luocontrolling} to explore their capability in handling such complex scenarios. We consider two common real-world scenarios with degraded images that contain 1) blur and noise or 2) haze and rain.
	For the noise and blur case, we combine the noisy images from BSD400 \cite{arbelaez2010contour} and WED \cite{ma2016waterloo} and blurry images from GoPro \cite{nah2017deep} for co-training. For the haze and rain case, we train models combining hazy images from SOTS \cite{ancuti2018haze} and rainy images from SPANet \cite{Wang_2019_CVPR}. For the evaluation, we collect 49 real images of haze and rain or blur and noise. The images and restored results are in the supplementary materials.
		\begin{table}[!h]
		\centering
		\vspace{-0.2cm}
		\caption{Quantitative comparison of the quality assessment of the 49 Real multiple-degradation images with average NIQE($\downarrow$)/PIQE($\downarrow$) metrics. }
		\vspace{-0.2cm}
		\label{multi-deg-num}
		\resizebox{0.38\textwidth}{!}{
				\begin{tabular}{lcc}
				\toprule
				Method & Blur and Noise & Rain and Haze\\
				\midrule
				Restormer \cite{Zamir2021Restormer} & 8.06/92.54&8.96/118.8\\
				IR-SDE \cite{luo2023image} &8.14/97.20 &7.87/102.1\\
				AirNet \cite{li2022all} & 8.23/102.4&8.85/113.2\\
				PromptIR \cite{potlapalli2023promptir} &5.01/68.59 &6.28/92.79\\
				DA-CLIP \cite{luocontrolling}&7.87/82.84&7.09/105.7\\
				\midrule
				RCOT \cite{tang2024residualconditioned} & 4.82/60.23&6.33/89.64\\
				DA-RCOT &\textbf{3.75}/\textbf{46.27}&\textbf{4.62}/\textbf{57.40}\\
				\bottomrule
				\end{tabular}}
	\end{table}
	
	Table \ref{multi-deg-num} reports the non-reference quantitative results on 49 real multi-degradation images, which validates the effectiveness of DA-RCOT in handling real-world multiple degradations in a single image. The visual examples are illustrated in Figure \ref{multi-deg}. For the noise and blur cases, DA-CLIP \cite{luocontrolling} restores images with remaining blur and noise and arouses significant artifacts around the edges. PromptIR \cite{potlapalli2023promptir} produces results with over-smoothed structures. For the haze and rain cases, DA-CLIP removes the haze and rain inadequately while PromptIR performs better at removing the rain but fails to address the haze. In comparison, DA-RCOT effectively removes the noise and blur or haze and rain, producing sharp images with well-preserved structural contents.  These results demonstrate the adaptability of our DA-RCOT in addressing multiple real-world degradations.  The advantages can be credited to the optimal restoration learning between multi-domain distributions, along with the conditioning of residual embeddings, which enables the OT map to adaptively adjust its behaviors for structure-preserving and degradation-aware restoration.
	\subsection{Ablation Studies}
	\subsubsection{Effect of residual embeddings conditioning (REC)}
	\begin{figure}[!h]
		\setlength\tabcolsep{1pt}
		\renewcommand{\arraystretch}{0.75} 
		\centering
		\begin{tabular}{ccc}
			\includegraphics[width=0.32\linewidth]{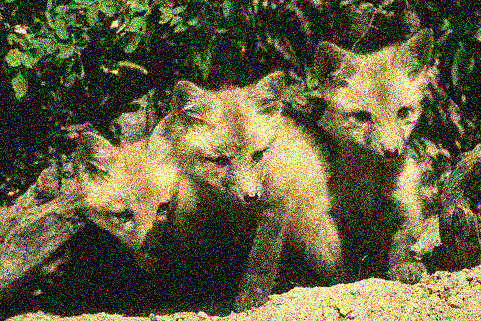}&
			\includegraphics[width=0.32\linewidth]{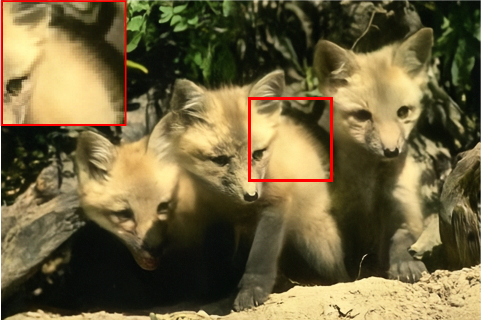}&
			\includegraphics[width=0.32\linewidth]{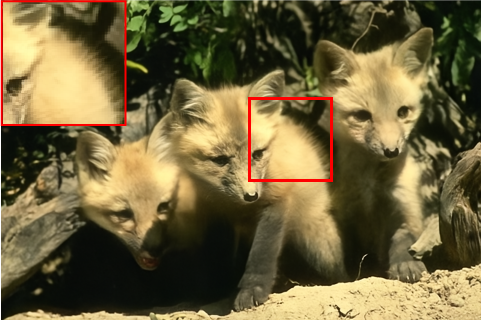}\\
			\includegraphics[width=0.32\linewidth]{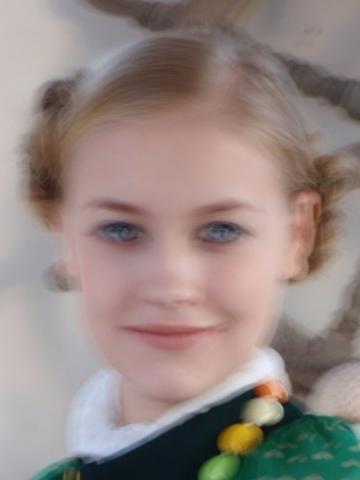}&
			\includegraphics[width=0.32\linewidth]{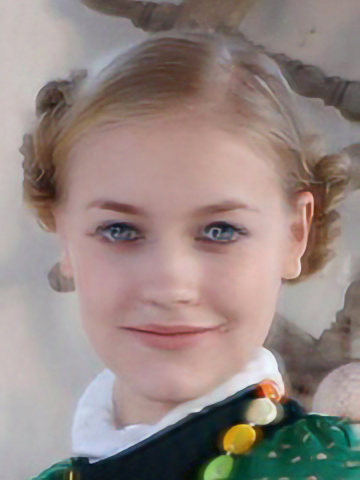}&
			\includegraphics[width=0.32\linewidth]{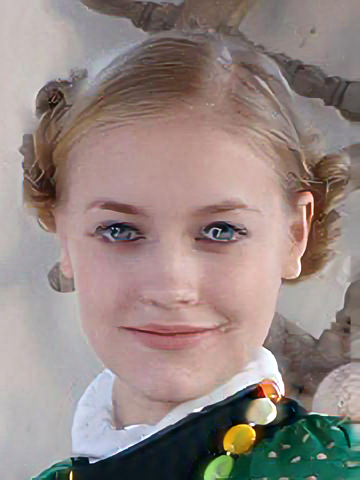}\\
			\includegraphics[width=0.32\linewidth]{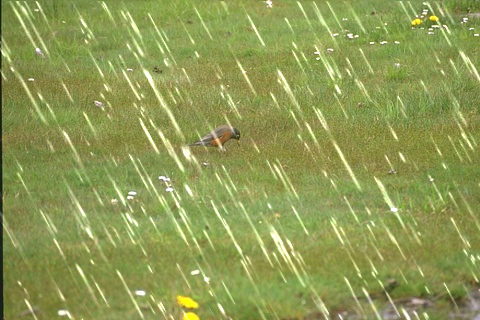}&
			\includegraphics[width=0.32\linewidth]{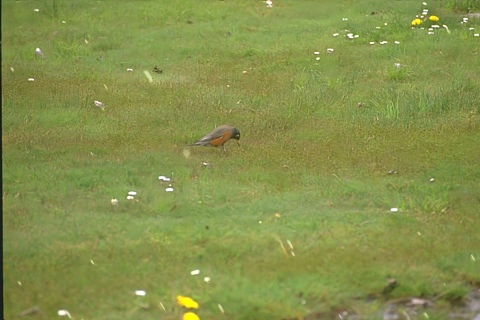}&
			\includegraphics[width=0.32\linewidth]{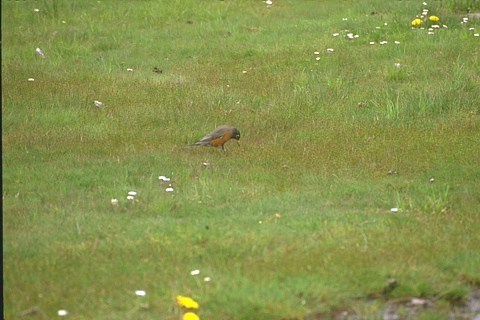}\\
			Degraded&w/o REC&w/ REC
		\end{tabular}
		\caption{Visual results to show the importance of REC. The model with REC produces sharper images with better structural contents.}
		\vspace{-0.3cm}
		\label{F-REC}
	\end{figure}
	We first justify the importance of the two-pass residual embeddings conditioning (REC) mechanism, i.e., utilizing the multi-scale residual embeddings $\{\mathbf R_i\}_{i=1,2,3}$ as conditions for the OT map. We try different conditions for the OT map: a) w/o any condition, b) intermediate restored result $\hat x_0$,  c) original residual embedding $\mathbf R_0$ (RCOT), and d) multi-scale residual embeddings $\{\mathbf R_i\}_{i=1,2,3}$ (DA-RCOT). 
	
		\begin{table}[!htbp]
		\centering
		\caption{Effect of using different losses. We report the average metrics of models using different losses for the three-degradation configuration.}
		\vspace{-0.2cm}
		\label{loss}
		\resizebox{0.48\textwidth}{!}{
			\begin{tabular}{c|l|cccc}
				\toprule
				&Loss&PSNR $\uparrow$&SSIM $\uparrow$&LPIPS $\downarrow$&FID $\downarrow$\\
				\midrule
				\multirow{2}{*}{Unpaired}&$\mathcal L_{u}$ &29.28&0.855&0.106&67.13\\
				&$\mathcal L_u+\mathcal L_{task}$ &30.20&0.897&0.086&53.24\\
				\midrule
				\multirow{3}{*}{Paired}& $L_1$ distance &31.82&0.911&0.087&53.21\\
				& $\mathcal L_p$ &32.35&0.915&0.063&37.12\\
				& $\mathcal L_p+\mathcal L_{task}$ &\textbf{32.60}&\textbf{0.917}&\textbf{0.055}&\textbf{30.06}\\
				\bottomrule
		\end{tabular}}
	\end{table}
	\begin{table*}[!t]
		\centering
		\vspace{-0.4cm}
		\caption{Results using transport costs with $g(\hat r)$ (FROT) and without $g(\hat r)$. Metrics are reported in the form of PSNR($\uparrow$)/LPIPS($\downarrow$).}
		\label{AFROT}
		\setlength{\tabcolsep}{12pt}
		\renewcommand{\arraystretch}{1}
		\resizebox{\textwidth}{!}{
			\begin{tabular}{l|ccccc|c}
				\toprule
				Method& SOTS &Rain100L &  BSD68\textsubscript{$\sigma$=25} & GoPro& LOL &Average  \\
				\midrule
				w/o $g(\hat r)$ &30.84/0.019 &37.28/0.018&31.02/0.096&28.24/0.158&22.87/0.103&30.05/0.079\\
				w/ $g(\hat r)$ & \textbf{31.26}/\textbf{0.007}&\textbf{38.22}/\textbf{0.008}&\textbf{31.22}/\textbf{0.080}&\textbf{28.86}/\textbf{0.135}&\textbf{23.25}/\textbf{0.084}&\textbf{30.40}/\textbf{0.064}\\
				\bottomrule
		\end{tabular}}
		\vspace{-0.3cm}
	\end{table*}
	Table \ref{T-REC} presents the All-in-One results on five degradation configurations. We can observe that the two-pass residual embedding conditioning mechanism provides a notable improvement as compared to the methods conditioned or nothing or $\hat x_0$, demonstrating that the residual embeddings capture valuable degradation semantics for AIR. As compared to the original $\mathbf R_0$, multi-scale residual embeddings $\{\mathbf R_i\}_{i=1,2,3}$ encode and decouple more intrinsic degradation knowledge and structural information, adapting the OT map for degradation-aware restoration. Figure \ref{F-REC} provides visual examples of denoising, deblurring, and deraining with no conditioning (w/o REC) and with $\{\mathbf R_i\}_{i=1,2,3}$ as conditions (w/ REC). The results show that the model without REC removes the degradations inadequately and causes loss of details, while the one with REC can restore sharp images with better structural details.

	\subsubsection{Effect of using different losses}
	We further explore the effect of using different losses to validate the effectiveness of FROT objectives (Eq. \ref{saddle} and \ref{pair-saddle}) for modeling AIR and the  $\mathcal L_{task}$ (Eq. \ref{task}) for enhancing the task semantic features of dense residual embedding $\mathbf R_1$.
	Table \ref{loss} reports the quantitative All-in-One results under the three-degradation configuration. The results show that in both unpaired and paired settings, incorporating $\mathcal{L}_{task}$ to emphasize the task semantics brings a significant improvement, particularly to the unpaired case.
	On the other hand, the FROT objective $\mathcal{L}_p$ outperforms the $L_1$ distance for supervised learning, with an average improvement of 0.53 dB in PSNR and a reduction of 0.024 in LPIPS. Notably, the perceptual quality metrics (LPIPS and FID) of $\mathcal L_u+\mathcal L_{task}$ for unpaired learning evenly match those of the $L_1$ distance case for paired learning. These results substantiate the advantage of modeling AIR from an optimal transport perspective with residual guidance. 
	
	\subsubsection{Effect of the residual regularization in FROT objective}
	We investigate the effect of the residual regularization in the FROT objective. In Table \ref{AFROT}, we compare the performance of the models trained under the FROT cost (w/ $g(\hat r)$) and OT cost (w/o $g(\hat r)$). The results show that residual regularization, which integrates degradation-specific knowledge into transport costs,  brings meaningful gains.
	\subsection{Discussion and Analysis}
	
	\subsubsection{Parameter quantity and computational complexity}
	We compare the parameter quantity and computational complexity in Table \ref{para}. We compare with the most recent methods IR-SDE \cite{luo2023image}, PromptIR \cite{potlapalli2023promptir}, and DA-CLIP \cite{luocontrolling}.  Table \ref{para} along with previous results shows that our DA-RCOT achieves significantly superior performance while incurring a slight increase in computational cost. The inference time and parameter efficiency of DA-RCOT are superior compared to the state-of-the-art method DA-CLIP \cite{luocontrolling}.
	\begin{table}[!h]
		\centering
		\vspace{-0.2cm}
		\caption{Comparison of the number of parameters, model computational efficiency, and inference time. The flops and inference time are computed on deraining images of size 256$\times$256.}
		\vspace{-0.2cm}
		\label{para}
		\resizebox{0.42\textwidth}{!}{
			\begin{tabular}{c|cccc}
				\toprule
				Method  & IR-SDE  & PromptIR & DA-CLIP&DA-RCOT \\
				\midrule
				\#Param&36.2M&36.3M&174.1M&49.8M\\
				Flops&117G&158G&118.5G&218G\\
				Inference time&4.23s&0.15s&4.59s&0.22s\\
				\bottomrule
		\end{tabular}}
		\vspace{-0.3cm}
	\end{table}
	
	\subsubsection{Generalization of REC beyond network designs} 
	\begin{table}[!h]
		\centering
		\caption{The influence of the multi-scale REC being integrated into different network backbones. The metrics are presented as PSNR ($\uparrow$)/SSIM ($\uparrow$)/LPIPS ($\downarrow$)/FID ($\downarrow$) values.}
		\label{trc}
		\setlength{\tabcolsep}{3pt}
		\resizebox{0.48\textwidth}{!}{
			\begin{tabular}{c|ccc}
				\toprule
				Method  & MPRNet \cite{Zamir2021MPRNet} & NAFNet \cite{chu2022nafssr} & Restormer \cite{Zamir2021Restormer} \\
				\midrule
				w/o REC & 34.95/0.964/0.039/21.61  & 35.58/0.969/0.036/18.35 & 36.74/0.978/0.023/13.29  \\
				\midrule
				w/ REC& 36.78/0.972/0.022/12.58&37.10/0.976/0.017/11.86&38.22/0.984/0.014/7.013 \\
				\bottomrule
		\end{tabular}}
	\end{table}
	The two-pass multi-scale REC mechanism forms a plug-in module (Figure \ref{model}), allowing us to use any architecture as a base model to generate the restored result, and then use this result to calculate the residual for the second-pass restoration. We have now included a comparison on Rain100L dataset \cite{fan2019general} between the MPRNet \cite{Zamir2021MPRNet}, NAFNet \cite{chu2022nafssr}, Restormer \cite{Zamir2021Restormer} methods and the corresponding methods with the proposed REC mechanism.
	The results in Table \ref{trc} show that the REC mechanism brings meaningful boosts to three benchmark network architectures, which validates its versatility and generalizability beyond network designs.
	
	\subsubsection{Training Cost Curves}
\begin{figure}[!h]
		\centering
		\includegraphics[width=1\linewidth]{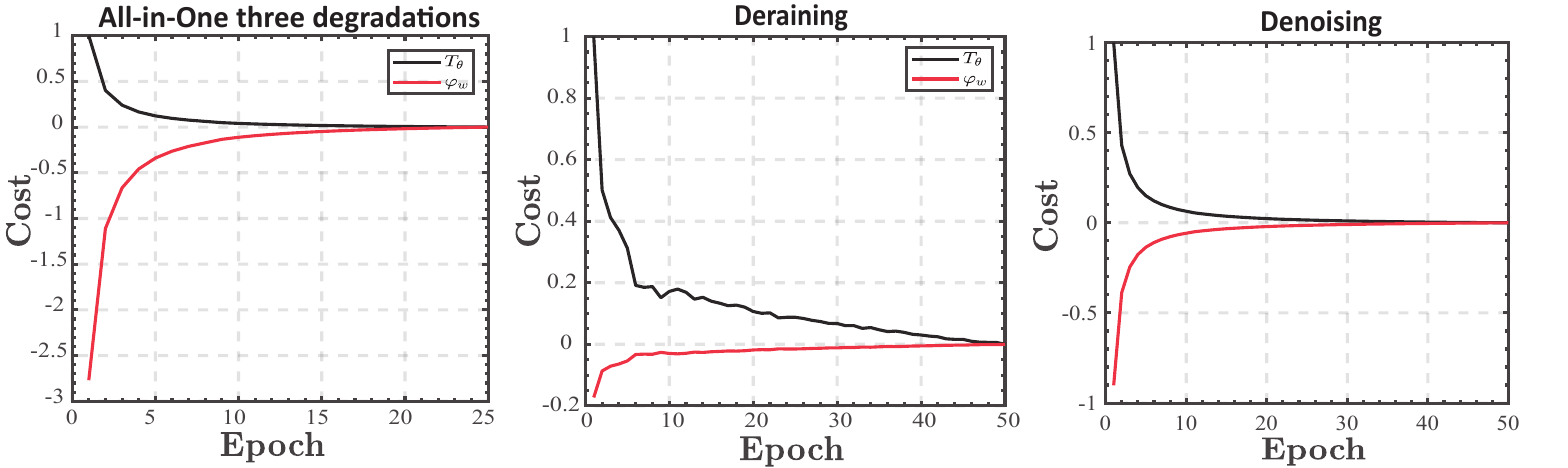}
		\vspace{-0.6cm}
		\caption{The cost curves for three tasks. The cost of $T_\theta$ is scaled to $[0,1]$. The cost of $\varphi_w$ is scaled to $[0,1]$ and then takes the negative.}
		\label{curve}
		
	\end{figure}

	In Figure \ref{curve}, we provide the cost curves over All-in-One three degradations and task-specific deraining and denoising of the transport map $T_\theta$ and the potential $\varphi_w$ in the training process. The $T_\theta$ cost curve is normalized to $[0,1]$. $\varphi_w$ cost is scaled to $[0,1]$ and then take the negative. 
	We can observe that the curves of the transport map and the potential converge well in an adversarial manner.
	\subsection{CONCLUSION, LIMITATION AND FUTURE WORK}
	This paper has proposed a Degradation-Aware Residual-Conditioned Optimal Transport (DA-RCOT) approach for the AIR problem, which models AIR as an OT problem and introduces the novel transport residual as a degradation-specific cue for both transport cost and transport map. We condition the transport map on the decoupled multi-scale residual embeddings via a two-pass process. This process injects the intrinsic degradation knowledge and structural information from multi-scale residual embedding into the OT map, which can thereby adaptively adjust its behaviors for structure-preserving all-in-one restoration. Extensive experiments demonstrate that DA-RCOT achieves state-of-the-art performance on a range of image restoration tasks under all-in-one and task-specific configurations. Particularly, DA-RCOT delivers superior adaptability to real-world scenarios even with multiple degradations and unique robustness to both degradation levels and the number of degradations.
	
	We acknowledge that there are some potential limitations. For example, when the noise level is much too severe, the generalization performance of DA-RCOT can deteriorate to some extent since the severe noise as a condition may interrupt the original features. In the future, we are interested in adapting the OT model for multi-modality one-to-many generation and translation problems.
	
	
	
	%
	
	\bibliographystyle{ieeetr}
	\bibliography{ref}

\end{document}